%% file: main.tex
\documentclass[10pt,twocolumn,letterpaper]{article}

\usepackage[pagenumbers]{cvpr} 

\usepackage{graphicx}
\usepackage{amsmath}
\usepackage{amssymb}
\usepackage{booktabs}
\usepackage{times}
\usepackage{epsfig}
\usepackage{graphicx}
\usepackage{amsmath}
\usepackage{amssymb}
\usepackage{placeins}
\usepackage{adjustbox}
\usepackage{multirow}
\usepackage{stackengine}
\usepackage{caption}
\usepackage{booktabs}
\input{symbols}

\usepackage{widetable}

\usepackage{color}
\usepackage[dvipsnames]{xcolor}

\usepackage{xspace}
\usepackage{float}
\usepackage{contour}
\usepackage{xkeyval}

\usepackage[pagebackref,breaklinks,colorlinks]{hyperref}

\usepackage[capitalize]{cleveref}
\crefname{section}{Sec.}{Secs.}
\Crefname{section}{Section}{Sections}
\Crefname{table}{Table}{Tables}
\crefname{table}{Tab.}{Tabs.}

\def\cvprPaperID{5427} 
\def\confName{CVPR}
\def\confYear{2022}

\newcommand{\ouralg}{MVS2D\xspace}
\newcommand\xrowht[2][0]{\addstackgap[.5\dimexpr#2\relax]{\vphantom{#1}}}
\newcommand{\myparagraph}[1]{\vspace{1pt} \noindent {\textbf{#1}}}

\newcommand{\wh}{\color{white}} 
\contourlength{.4pt}
\newcommand{\figlabel}[1]{\sffamily\bfseries\wh\scriptsize\contour{black}{#1}} 
\makeatletter
\newlength{\sfp@hseplen}\newlength{\sfp@vseplen}
\define@cmdkey{subfigpos}[sfp@]{vsep}[0.6\baselineskip]{\setlength{\sfp@vseplen}{\sfp@vsep}}
\define@cmdkey{subfigpos}[sfp@]{hsep}[2.5pt]{\setlength{\sfp@hseplen}{\sfp@hsep}}
\newcommand{\subfigimg}[3][,]{%
  \setkeys{Gin,subfigpos}{vsep,hsep,#1}
  \setbox1=\hbox{\includegraphics{#3}}
  \leavevmode\rlap{\usebox1}
  \rlap{\hspace*{\sfp@hsep}\raisebox{\dimexpr\ht1-6pt}{\figlabel{#2}}}
  \phantom{\usebox1}
}
\makeatother

\begin{document}

\title{\ouralg: Efficient Multi-view Stereo via Attention-Driven 2D Convolutions}

\author{
  \hspace{-1.3cm}
  \begin{tabular}[t]{c}
    Zhenpei Yang$^{1,*}$ \quad  Zhile Ren$^2$ \quad Qi Shan$^2$ \quad Qixing Huang$^{1}$\\
    $^1$The University of Texas at Austin \quad $^2$Apple\\
\end{tabular}
}
\maketitle
\let\thefootnote\relax\footnotetext{$^*$ Experiments are conducted by Z. Yang at The University of Texas at Austin. Email: yzp@utexas.edu}

\input{00_abstract.tex}
\input{01_intro.tex}

\input{02_related.tex}
\input{03_approach.tex}

\input{04_results.tex}

\input{05_conclusions.tex}

{\small
\bibliographystyle{ieee_fullname}
\bibliography{refs}
}

\end{document}

%% file: symbols.tex
\let \bs = \boldsymbol
\let \set = \mathcal

\newcommand{\R}{\mathcal{R}}

%% file: 00_abstract.tex
\begin{abstract}
Deep learning has made significant impacts on multi-view stereo systems. State-of-the-art approaches typically involve building a cost volume, followed by multiple 3D convolution operations to recover the input image's pixel-wise depth. While such end-to-end learning of plane-sweeping stereo advances public benchmarks' accuracy, they are typically very slow to compute. 
We present \ouralg, a highly efficient multi-view stereo algorithm that seamlessly integrates multi-view constraints into single-view networks via an attention mechanism. Since \ouralg only builds on 2D convolutions, it is at least $2\times$ faster than all the notable counterparts. Moreover, our algorithm produces precise depth estimations and 3D reconstructions, achieving state-of-the-art results on challenging benchmarks ScanNet, SUN3D, RGBD, and the classical DTU dataset. our algorithm 
also out-performs all other algorithms in the setting of inexact camera poses. Our code is released at \url{https://github.com/zhenpeiyang/MVS2D}
\end{abstract}

%% file: 01_intro.tex
\section{Introduction}
Multi-view Stereo (MVS) aims to reconstruct the underlying 3D scene or estimate the dense depth map using multiple neighboring views. It plays a key role in a variety of 3D vision tasks. With high-quality cameras becoming more and more accessible, there are growing interests in developing reliable and efficient stereo algorithms in various applications, such as 3D reconstruction, augmented reality, and autonomous driving. As a fundamental problem in computer vision, MVS has been extensively studied~\cite{Furukawa:2015:MST}. Recent research shows that deep neural networks, especially convolutional neural networks (CNNs), lead to more accurate and robust systems than traditional solutions. Several approaches~\cite{kusupati2019normal,yu2020fast} report exceptional accuracy on challenging benchmarks like ScanNet~\cite{scannet17cvpr} and SUN3D~\cite{xiao2013sun3d}. 

State-of-the-art CNN-based multi-view approaches typically fall into three categories:   1) Variants of a standard 2D UNet architecture with feature correlation~\cite{mayer2016large,liang2018learning}. However, these approaches work best for rectified stereo pairs, and extending them to multi-view is nontrivial. 
2) Constructing a differential 3D cost volume~\cite{huang18deepmvs,im2019dpsnet,yao2018mvsnet,nie2019multi,yao2019recurrent,gu2020cascade}. These algorithms significantly improve the accuracy of MVS, but at the cost of heavy computational burdens. Furthermore, the predicted depth map by 3D convolution usually contains salient artifacts which have to be rectified by a 2D refinement network~\cite{im2019dpsnet}. 3) Maintain a global scene representation and fuse multi-view information through ray-casting features from 2D images~\cite{murez2020atlas}. This paradigm cannot handle large-scale scenes because of the vast memory consumption on maintaining a global representation.

\begin{figure}
    \includegraphics[width=0.85\linewidth]{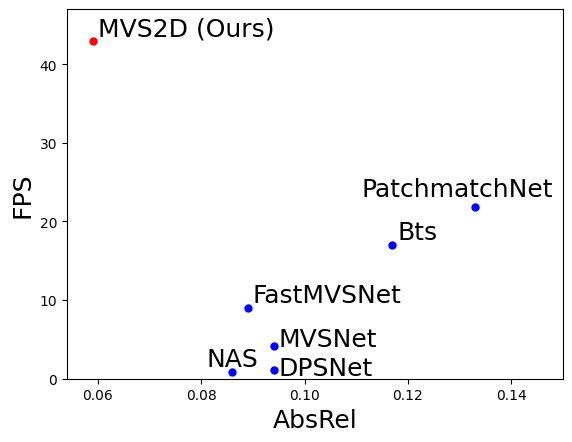}
\vspace{-0.1in}
\caption{Inference frame per second (FPS) vs.\ depth error (AbsRel) on ScanNet~\cite{scannet17cvpr}. Our model achieve significant reduction in inference time, while maintaining state-of-the-art accuracy. }
\label{fig:teaser}
\vspace{-0.1in}
\end{figure}

Aside from multi-view depth estimation, we have also witnessed the tremendous growth of single-view depth prediction networks~\cite{lee2019big,yin2019enforcing,xu2017multi,qi2018geonet,Yang_2021_CVPR}. As shown in Table \ref{table:scannet}, Bts~\cite{lee2019big} has achieved impressive result on ScanNet~\cite{scannet17cvpr}. Single-view depth prediction roots in learning feature representations to capture image semantics, which is orthogonal to correspondence-computation in multi-view techniques. A natural question is how to combine single-view depth cues and multi-view depth cues. 

We introduce \ouralg that combines the strength of single-view and multi-view depth estimations. The core contribution is an attention mechanism that aggregates features along epipolar lines of each query pixel on the reference images. This module captures rich signals from the reference images. Most importantly, it can be easily integrated into standard CNN architectures defined on the input image, introducing relatively low computational cost. 

Our attention mechanism possesses two appealing characteristics: 1) Our network only contains 2D convolutions. 2) Besides relying on the expressive power of 2D CNNs, the network seamlessly integrates single-view feature representations and multi-view feature representations. Consequently, \ouralg is the most efficient approach compared to state-of-the-art algorithms (See Figure~\ref{fig:teaser}). It is $48\times$ faster than NAS~\cite{kusupati2019normal}, $39\times$ faster than DPSNet~\cite{im2019dpsnet}, $10\times$ faster than MVSNet~\cite{yao2018mvsnet}, $4.7\times$ faster than FastMVSNet~\cite{yu2020fast}, and almost $2\times$ speed-up over the most recent fastest approach PatchmatchNet~\cite{wang2021patchmatchnet}. In the mean-time, \ouralg achieves state-of-the-art accuracy. 

Intuitively, the benefit of \ouralg comes from the early fusion of the intermediate feature representations. The outcome is that the intermediate feature representations contain rich 3D signals.
Furthermore, \ouralg offers ample space where we can design locations of the attention modules to address different inputs. One example is when the input camera poses are inaccurate, and corresponding pixels deviate from the epipolar lines on the input reference images. We demonstrate a simple solution, which installs multi-scale attention modules on an encoder-decoder network. In this configuration, corresponding pixels in down-sampled reference images lie closer to the epipolar lines, and \ouralg detect and rectify correspondences automatically.

We conduct extensive experiments on challenging benchmarks ScanNet~\cite{scannet17cvpr}, SUN3D~\cite{xiao2013sun3d}, RGBD~\cite{sturm12iros} and Scenes11~\cite{sturm12iros}. \ouralg  achieves the state-of-the-art performance on nearly all the metrics. Qualitatively, compared to recent approaches~\cite{im2019dpsnet,yao2018mvsnet,kusupati2019normal,yu2020fast}, \ouralg helps generate higher quality 3D reconstruction outputs.

%% file: 02_related.tex
\section{Related Works}
\label{Section:Related:Works}
\begin{figure*}[ht!]
\centering
\includegraphics[width=0.8\linewidth]{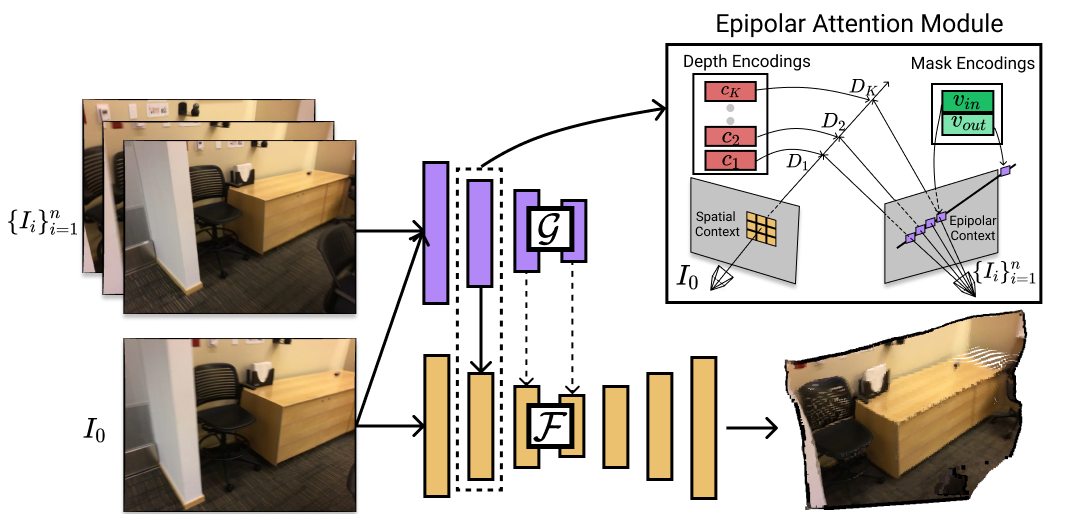}
\vspace{-0.1in}
\caption{Network architecture of \ouralg. We employ a 2D UNet structure $\mathcal{F}$ to make the depth prediction on $I_0$, while injecting multi-view cues extracted using $\mathcal{G}$ through the Epipolar Attention Module. Dashed arrows only exist in \textbf{Ours-robust} model (Section \ref{sec3.4}). We highlight that the proposed epipolar attention module can be easily integrated into most 2D CNNs. }
\label{fig:network}
\vspace{-0.1in}
\end{figure*}

\myparagraph{Recent advances of multi-view stereo.} 
Multi-view stereo algorithms can be categorized into depth map-based approaches, where the output is a per-view depth map, or point-based approaches, where the output is a sparse reconstruction of the underlying scene (cf.~\cite{Furukawa:2015:MST}). Many traditional multi-view stereo algorithms follow a match-then-reconstruct paradigm~\cite{furukawa2009accurate} that leverages the sparse-nature of feature correspondences. Such a paradigm typically fails to reconstruct textureless regions where correspondences are not well-defined. Along this line, Zbontar~\etal~\cite{zbontar2015computing} provided one of the first attempts to bring the power of feature learning into multi-view stereo. They proposed a supervised feature learning approach to find the correspondences. 
Recently, researchers have found that depth map-based approaches~\cite{kar2017learning,huang18deepmvs,yao2018mvsnet,im2019dpsnet,yao2019recurrent} are more favorable than those that follow the match-and-reconstruct paradigm. A key advantage of these approaches is that they can utilize the efficiency of regular tensor operations. \cite{yao2018mvsnet,im2019dpsnet} proposed an end-to-end plane-sweeping stereo approach that constructs learnable 3D cost volume. While MVSNet~\cite{yao2018mvsnet} focuses on the reconstruction of 3D scene, DPSNet~\cite{im2019dpsnet} focuses on evaluating the per-view depth-map accuracy. Researchers have also explored other 3D representations to regularized the prediction, such as point clouds~\cite{chen2019point}, surface normals~
\cite{kusupati2019normal}, or meshes~\cite{wang2020mesh}. There are also several benchmark datasets for this task~\cite{yao2020blendedmvs,xiao2013sun3d,sturm12iros,aanaes2016large,scannet17cvpr,ummenhofer17demon}.

\myparagraph{Cost volume for multi-view stereo.} A recent line of works on multi-view stereo utilizes the notion of \textsl{cost volume}, which contains feature matching costs for a pair of images~\cite{hosni2012fast}. This feature representation has been successfully implemented in various pixel-wise matching tasks like optical flow~\cite{sun2018pwc}. Authors of MVSNet~\cite{yao2018mvsnet} and DPSNet~\cite{im2019dpsnet} proposed to first construct a differentiable cost volume and then use the power of 3D CNNs to regularize the cost volume before predicting per-pixel depth or disparity. 
Most recent state-of-the-art approaches follow such a paradigm~\cite{yao2019recurrent,nie2019multi, gu2020cascade, chen2019point, yu2020fast,long2020multiview}. However, the size of the cost volume ($C\times K\times H\times W$) is linearly related to the number of depth hypotheses $K$. These approaches are typically  slow in both training and inference. For example, DPSNet~\cite{im2019dpsnet} takes several days to train on ScanNet; NAS~\cite{kusupati2019normal} takes even longer because of its extra training of a depth-normal consistency module. Recently, Murez~\etal~\cite{murez2020atlas} proposed to construct a volumetric scene representation from a calibrated image sequence for scene reconstruction. However, their approach is very memory demanding due to the high memory requirement of global volumetric representations.

\myparagraph{Efficient multi-view stereo.} 
Several recent works aim at reducing the cost of constructing cost volumes. Duggal~\etal~\cite{duggal2019deeppruner} prune the disparity search range during cost volume construction. Xu~\etal~\cite{xu2020aanet} integrate adaptive sampling and deformable convolution into correlation-based methods~\cite{mayer2016large,liang2018learning} to achieve efficient aggregation. Several other works~\cite{yao2019recurrent,tankovich2020hitnet,gu2020cascade} employ iterative refinement procedures. The above approaches either only work for pairwise rectified stereo matching tasks or have to construct a 3D cost-volume. Alternatively, Poms~\etal~\cite{poms2018learning} learn how to merge patch features for 3D reconstruction efficiently. Badki~\etal~\cite{badki2020bi3d} convert depth estimation as a classification task, but the resulting accuracy is not state-of-the-art. Recently, Yu~\etal~\cite{yu2020fast} proposed constructing a sparse cost-volume through regular sub-sampling and then applying Gauss-Newton iterations to refine the dense depth map. Wang~\etal~\cite{wang2021patchmatchnet} proposed a highly-efficient Patchmatch-inspired approach for MVS tasks. In contrast, we take an orthogonal approach based on the attention-driven 2D convolutions.

\myparagraph{Attention in 3D vision}
Attention mechanism has shown prominent results on both natural language processing (NLP) tasks~\cite{vaswani2017attention} and vision tasks~\cite{wang2018non}. Recently, self-local attention~\cite{ramachandran2019stand, shaw2018self} has shown promising results compared with the convolution-based counterparts. 
Several recent works that build an attention mechanism in MVS~\cite{luo2020attention,zhang2021long,long2020multiview}, but still rely on 3D CNNs and cannot avoid constructing a heavy-weighted cost volume. A promising direction is to utilize a geometry-aware 2D attention mechanism. Recent works have shown that this paradigm works well for active sensing~\cite{cheng2018geometry,tung2019learning} and neural rendering~\cite{tobin2019geometry}. Motivated by these works, we propose an epipolar attention module in this paper. The key contribution is a network design that aggregates single-view depth cues and multi-view depth cues to output accurate MVS outputs.

%% file: 03_approach.tex
\section{Approach}
We provide an overview of the network architecture of \ouralg in Fig.~\ref{fig:network}. We operate in a multi-view stereo setting (Sec.~\ref{sec3.1}), and employ a 2D UNet structure in our network design (Sec.~\ref{sec3.2}). Our core contribution is the epipolar attention module (Sec.~\ref{sec3.3}--\ref{sec3.4}), which is highly accurate and efficient (Sec.~\ref{sec3.5}) for depth estimation (Sec.~\ref{sec3.6}).

\subsection{Problem Setup}
\label{sec3.1}

We aim to estimate the per-pixel depth for a source image $I_0 \in \R^{h\times w \times 3}$, given $n$ reference images $\{I_i\}_{i=1}^{n}$ of the same size captured at nearby views.
We assume the source image and the reference images share the same intrinsic camera matrix $\set{K} \in \R^{3\times3}$, which is given. We also assume we have a good approximation of the relative camera pose between the source image and each reference image $T_i = (R_i|\bs{t}_i)$
,where $R_i\in\text{SO}(3)$ and $\bs{t}_i\in \R^3$. $T_i$ usually comes from the output of a multi-view structure-from-motion algorithm. Our goal is to recover the dense pixel-wise depth map associated with $I_0$. 

We denote the homogeneous coordinate of a pixel $p_0$ in the source image $I_0$ as $\overline{\bs{p}}_0 = \begin{pmatrix}
\overline{p}_{0,1},\overline{p}_{0,2},1
\end{pmatrix}^T$. Given the depth $d_0\in \R$ of $p_0$, the unprojected 3D point of $p_0$ is 
$$
\bs{p}_0(d_0) = d_0\cdot (\set{K}^{-1}\bs{p}_0).
$$

Similarly, we use $\bs{p}_i(d_0)$ and $\overline{\bs{p}}_i(d_0)$ to denote respectively the 3D coordinates and homogeneous coordinates of $\bs{p}_0(d_0)$ in the $i$-th image's coordinate system. They satisfy
\begin{align}
\bs{p}_i(d_0) &= R_i\bs{p}_0(d_0) + \bs{t}_i, \nonumber \\
\overline{\bs{p}}_i(d_0) &= \mathcal{K}\bs{p}_i(d_0).
\label{eq1}
\end{align}

\subsection{Network Design Overview}
\label{sec3.2}

In this paper, we innovate developing a multi-view stereo approach that only requires 2D convolutions. Specifically, similar to most single-view depth prediction networks, our approach progressively computes multi-scale activation maps of the source image and outputs a single depth map. The difference is that certain intermediate activation maps combine both the output of a 2D convolution operator applied to the previous activation map and the output of an attention module that aggregates multi-view depth cues. This attention module, which is the main contribution of this paper, matches each pixel of the source image and corresponding pixels on epipolar lines on the reference images. The matching procedure utilizes learned feature activations on both the source image and the reference images. The output is encoded using learned depth codes compatible with the activation maps of the source image. 

Formally speaking, our goal is to learn a feed-forward network $\set{F}$ with $L$ layers. With $\set{F}_j\in \R^{h_j\times w_j \times m_j}$ we denote the output of the $j$-th layer, where $m_j$ is its feature dimension, $h_j$ and $w_j$ are its height and width. Note that the first layer $\set{F}_1 \in \R^{h_1\times w_1\times 3}$ denotes the input, while the last layer $\set{F}_{L}\in \R^{h_L\times w_L}$ denotes the output layer containing depth prediction. Between two consecutive layers are a general convolution operator $C_j: \R^{h_j\times w_j\times m_j}\rightarrow \R^{h_j\times w_j\times m_{j+1}}$ (it can incorporate standard operators such as down-sampling, up-sampling,  and max-pooling) and an optional attention module $\set{A}_j:\R^{h_j\times w_j\times m_j}\rightarrow \R^{h_j\times w_j\times m_j}$:
$$
\set{F}_{j+1} = \set{C}_j\circ \set{A}_j \circ \set{F}_j.
$$
As we will see immediately, the attention operator $\set{A}_j$ utilizes features extracted from the reference images. Without these attention operations, $\set{F}$ becomes a standard encoder-decoder network for single-view depth prediction. 

Another characteristic of this network design is that the convolution operator $\set{C}_j$ implicitly aggregates multi-view depth cues extracted at adjacent pixels. This approach promotes consistent correspondences among adjacent pixels that share the same epipolar line or have adjacent epipolar lines.

\subsection{Epipolar Attention Module}
\label{sec3.3}

We proceed to define $\set{A}_j(p_0)$, which is the action of $\set{A}_j$ on each pixel $p_0$. It consists of two parts:
\begin{equation}
\set{A}_j(p_0) = \set{A}_{j}^{\text{ep}}(p_0, \{I_i\}_{i=1}^{n}) + \set{A}_j^{0}(\set{F}_j(p_0)) .
\label{Eq:Attention0}
\end{equation}
As we will define next, $\set{A}_{j}^{\text{ep}}(p_0, \{I_i\}_{i=1}^{n})$ uses trainable depth codes to encode the matching result between $p_0$ and the reference images. $\set{A}_j^{0}:\R^{m_j}\rightarrow \R^{m_j}$ is composed of an identity map and a trainable linear map that transforms the feature associated with $p_0$ in $\set{F}_j$. 

The formulation of $\set{A}_{j}^{\text{ep}}(p_0, \{I_i\}_{i=1}^{n})$ uses the \textsl{epipolar context} of $p_0$. It consists of samples on the epipolar lines of $p_0$ on the reference images. These samples are obtained from sampling the depth values $d_0$ of $p_0$ and then applying (\ref{eq1}). With $p_{i}^{k}$ we denote the $k$-th sample on the $i$-th reference image.

To match $p_0$ and $p_i^{k}$, we introduce a feature extraction network $\set{G}$ that has identical architecture (except the attention modules) as $\set{F}_{j_{\max}}$ where $j_{\max}$ is the maximum depth of any attention module of $\set{F}$. With $\set{G}_j(I_0,p_0)\in \R^{m_j}$ and $\set{G}_j(I_i, p_i^k)\in \R^{m_j}$ we denote the extracted features of $p_0$ and $p_i^k$, respectively.
Following the practice of scaled-dot product attention~\cite{vaswani2017attention}, we introduce two additional trainable linear maps $\bs{f}_{0}^j: \R^{m_j}\rightarrow \R^{m_j}$ and $\bs{f}_{\text{ref}}^j: \R^{m_j}\rightarrow \R^{m_j}$ to transform the extracted features. With this setup, we define the matching score between $p_0$ and $p_i^{k}$ as 
\begin{equation}
w_{ik}^j = \big(\bs{f}_{0}^j(\set{G}_j(I_0,p_0))\big)^T\big(\bs{f}_{\text{ref}}^j(\set{G}_j(I_i, p_i^k))\big).
\label{Eq:Matching:Weight}
\end{equation}

It remains to 1) model samples that are occluded in the reference images, and 2) bridge the weights $w_{ik}^j$ defined in (\ref{Eq:Matching:Weight}) and the input to the convolution operator $\set{C}_j$. To this end, we first introduce trainable mask codes $\bs{c}_{jk}\in \R^{m_j}$ that correspond to the $k$-th depth sample. We then introduce $\bs{v}_{\text{in}}^j\in \R^{m_j}$ and $\bs{v}_{\text{out}}^j\in \R^{m_j}$, which are trainable codes for inside and outside samples, respectively. Define
\begin{equation}
\bs{v}_{ik}^{j} = \left\{
\begin{array}{cc}
\bs{v}_{\text{in}}^j & 0\le \overline{p}_{i,1}^k< w, 0\le \overline{p}_{i,2}^k< h, p_{i,3}^k \ge 0,\\
\bs{v}_{\text{out}}^j & \textup{otherwise}
\end{array}
\right.\
\label{Eq:Mask:Codes}
\end{equation}
where $\overline{p}_{i}^k = (\overline{p}_{i,1}^k,\overline{p}_{i,2}^k,1)^T$, $p_{i}^k = (p_{i,1}^k,p_{i,2}^k,p_{i,3}^k)^T$.
To enhance the expressive power of $\set{G}_j$, we further include a trainable linear map $\set{A}_j^{1}$ that depends only on feature of $p_0$ and not on the matching results. Combing with (\ref{Eq:Matching:Weight}) and (\ref{Eq:Mask:Codes}), we define 
\begin{equation}
\set{A}_{j}^{\text{ep}}(p_0, \{I_i\}_{i=1}^{n}) =\set{A}_j^{1}(\set{G}_j(p_0))+\sum\limits_{i=1}^{n}\sum\limits_{k=1}^{K} \mathcal{N}(\frac{w_{ik}^j}{\sqrt{m_j}}) (\bs{v}_{ik}^j\odot \bs{c}_k))
\label{Eq:Attention:2}    
\end{equation}
where $\mathcal{N}$ is the softmax normalizing function over $\frac{w_{ik}^j}{\sqrt{m_j}}, 1\leq k \leq K$.
Substituting (\ref{Eq:Attention:2}) into (\ref{Eq:Attention0}), the final attention module is given by 
\begin{align}
    \set{A}_j(p_0) &= \set{A}_j^{0}(\set{F}_j(p_0)) + \set{A}_j^{1}(\set{G}_j(p_0))\nonumber\\ &+\sum\limits_{i=1}^{n}\sum\limits_{k=1}^{K} \mathcal{N}(\frac{w_{ik}^j}{\sqrt{m_j}})\nonumber (\bs{v}_{ik}^j\odot \bs{c}_k)).
\end{align}

\begin{figure}[h]
   \begin{center}
    \includegraphics[width=0.4\textwidth]{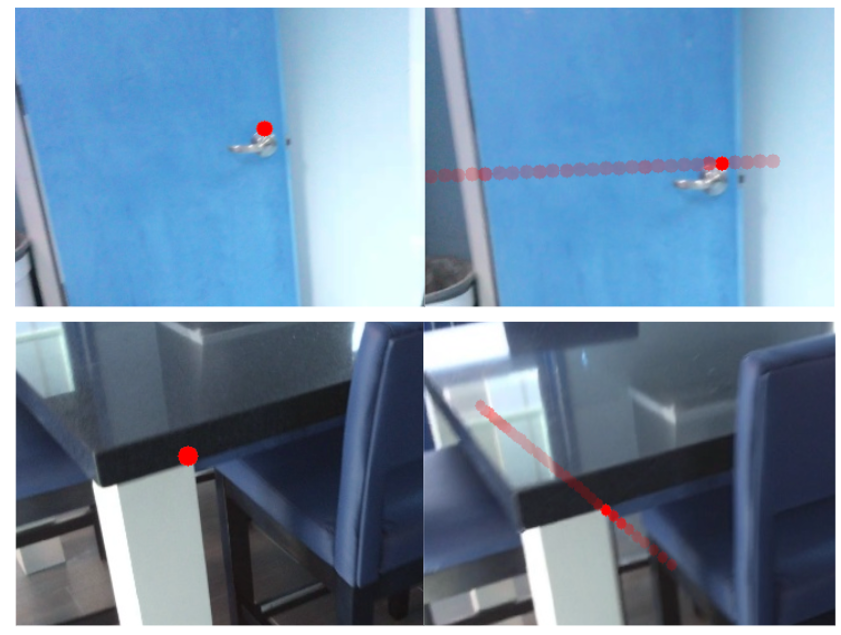}
  \end{center}
    \vspace{-0.2in}
  \caption{Visualization of attention scores. Left: source view with query pixels. Right: reference view with candidate pixels, where opacity is learned attention scores.}
  \label{fig:visualization_attention}
\end{figure}

Note that the attention modules at different layers have different weights.  Eq.~\ref{Eq:Matching:Weight} can be viewed as a similarity score between source pixel and correspondence candidates. In Fig.~\ref{fig:visualization_attention}, we visualize the learned attention scores for query pixels. The true corresponding pixels on reference images have larger learned weights along epipolar lines. 

\subsection{Attention Design for Robust Multi-View Stereo}
\label{sec3.4}

Since the attention module assumes that the corresponding pixels lie on the epipolar lines, the accuracy of \ouralg depends on the relative poses' accuracy between the reference images and the source image. When input poses are accurate, our experiments suggest a single attention module at the second layer of $\set{F}$ is sufficient. This leads to a highly efficient multi-view stereo network. 

When the input poses are inexact, we address this issue by installing attention modules at different resolutions of the input images, i.e., at different layers of $\set{F}$. This approach ensures that the corresponding pixels lie sufficiently close to the epipolar lines at those resolutions at coarse resolutions. Since the convolution operations $\set{C}_j$ at different layers of $\set{F}$ aggregate multi-view features extracted at different pixels, we find this simple network design implicitly rectifies inexact epipolar lines. Figure~\ref{fig:network} illustrates the attention modules under these two cases.

\begin{table}[!t]
\centering
\resizebox{\linewidth}{!}{%
\begin{tabular}{rccccc}
\toprule
\multicolumn{1}{c}{Method}   & FPS (3)$\uparrow$  & FPS(7)$\uparrow$  & FPS(11) $\uparrow$& Param (M) $\downarrow$ & AbsRel$\downarrow$  \\ \hline
Bts~\cite{lee2019big} & 17.0 & - & - & 46.8 & 0.117\\ 
\hline
MVSNet~\cite{yao2018mvsnet}  & 4.1 & 2.4 & 1.6 & 1.1 & 0.094\\
DPSNet~\cite{im2019dpsnet}  & 1.1 & 0.7 &0.5  & 4.2& 0.094\\
FastMVS~\cite{yu2020fast}& 9.0 & 6.0 & 4.3 & 0.4 &  0.089\\
PatchmatchNet~\cite{wang2021patchmatchnet}  & 21.8  & 11.6 & 8.5 &\textbf{0.2} &  0.133\\
NAS~\cite{kusupati2019normal}  & 0.9  & 0.6 & 0.4 &18.0 &  0.086\\
\hline
Ours-mono  & \textbf{94.7} & - & - & 12.3 & 0.145\\ 
Ours-robust  & 17.5 & 10.1 & 7.1& 24.4 & \textbf{0.059}\\ 
Ours  & 42.9 & \textbf{29.1} & \textbf{21.8} & 13.0 & \textbf{0.059}\\ 

\bottomrule
\end{tabular}
}
\caption{Quantitative comparison on computational efficiency. FPS ($V$) only applies to multi-view methods~\cite{yao2018mvsnet,yu2020fast,im2019dpsnet,kusupati2019normal} and means we use $V$ images to make the prediction. Note that numbers under the AbsRel metric are identical to those in Table~\ref{table:scannet} for ease of comparison. We use a single Nvidia V100 GPU for measuring FPS. Please refer to section~\ref{sec4.4} for additional discussions.}
\label{table:speed}
\vspace{-0.1in}
\end{table}

\subsection{Computational Complexity}\label{sec3.5} 
For the sake of simplicity in notation, we assume the feature channel dimension $C$ is the same in both input and output. We denote the feature height and width as $H$ and $W$ respectively and denote the kernel size of convolution layers as $k$. Suppose there are $K$ depth samples,  the complexity of 3D convolution is $\mathcal{O}(C^2HWKk^3)$.

For our approach, the computational complexity for executing one layer of $\set{C}\circ \set{A}$ is in total $\mathcal{O}(CHW(Ck^2+K))$. Since $K$ is usually less than $Ck^2$, our module leads to a $Kk$ times reduction in computation. The actual runtime can be found in  Table~\ref{table:speed}.

\subsection{Training Details}\label{sec3.6}

Our implementation is based on Pytorch. For ScanNet and DeMoN, we simply optimize the $L_1$ loss between predicted and ground truth depth. For DTU, we introduce a simple modification, as was done in ~\cite{kendall2017uncertainties}, to simultaneously train a confidence prediction. We use Adam~\cite{DBLP:journals/corr/KingmaB14} optimizer with $\epsilon =10^{-8}$, $\beta = (0.9,0.999)$. We use a starting learning rate $2e^{-4}$ for ScanNet, $8e^{-4}$ for DeMoN and $2e^{-4}$ for DTU. Please refer to supp. material for more training details.

%% file: 04_results.tex
\section{Experimental Results}
\label{Section:Results}
\subsection{Datasets}\label{sec4.1}
\myparagraph{ScanNet~\cite{scannet17cvpr}} The ScanNet dataset contains 807 unique scenes with image sequences captured from different camera trajectories. We sample 86324 triple images (one source image and two reference images) for training and 666 triple images for testing. Our setup ensures the scene corresponding to test images is not included in the training set. 

\myparagraph{DeMoN~\cite{ummenhofer17demon}}
We further validate our method on DeMoN, which is a dataset introduced by~\cite{ummenhofer17demon} for multi-view depth estimation. The training set consists of three data sources, SUN3D~\cite{xiao2013sun3d}, RGBD~\cite{sturm12iros}, and Scenes11~\cite{ummenhofer17demon}. SUN3D and RGBD contain real indoor scenes, while Scenes11  is synthetic. In total, there are 79577 training pairs for SUN3D, 16786 for RGBD, and 71820 for Scenes11. 

\myparagraph{DTU~\cite{aanaes2016large}} While our approach is designed for multi-view depth estimation, we additionally validate our method on the DTU dataset, which has been considered as one of the main test-bed for multi-view reconstruction algorithms.

\subsection{Evaluation Metrics}\label{sec4.2}
\myparagraph{Efficiency.} We benchmark our methods against baseline methods on the frame per second~(FPS) during inference. We additionally compare the FPS when increasing the number of reference views. 

\myparagraph{Depth Accuracy.} We use the conventional metrics of depth estimation~\cite{lee2019big} (See Table~\ref{table:metric}). 
Note that in contrast to monocular depth estimation evaluation, we do not factor out the depth scale before evaluation. The ability to correctly predict scale will render our method more applicable. 

\myparagraph{Scene Reconstruction Quality.} We further apply \ouralg for scene reconstruction. We follow PatchmatchNet~\cite{wang2021patchmatchnet} to fuse the per-view depth map into a consistent 3D model. Please refer to supp. material for quantitative and more qualitative comparisons.

\myparagraph{Robustness under Noisy Input Pose.} We perturb the input relative poses $T_j$ during training and report the model performance on ScanNet test set in Table \ref{table:perturb}. Please refer to the supp. material for details of the pose perturbing procedures.

\begin{table}[!h]
\centering
\resizebox{\linewidth}{!}{%
\begin{tabular}{cc|cc}
\toprule
AbsRel & $\frac{1}{N}\sum_{i}\frac{|d_i - d_i^{*}|}{d_i^{*}}$ & RMSE & $\sqrt{\frac{1}{N}\sum_{i}(d_i - d_i^{*})^2}$ \\
\xrowht{10pt}
SqRel & $\frac{1}{N}\sum_{i}\frac{(d_i - d_i^{*})^2}{ d_i^{*}}$ & RMSELog & $\sqrt{\frac{1}{N}\sum_{i}(\log{d_i} - \log{d_i^{*}})^2}$\\
\xrowht{10pt}
AbsDiff & $\sqrt{\frac{1}{N}\sum_{i}|d_i - d_i^{*}|}$ & Log10 & $\frac{1}{N}\sum_{i}|\log_{10}d_i - \log_{10}d_i^{*}|$\\
\xrowht{10pt}
$\delta < 1.25^{k}$ & $\frac{1}{N}\sum_{i}(\max(\frac{d_i}{d_i^{*}},\frac{d_i^*}{d_i} ) < 1.25^{k})$ & thre@x & $\frac{1}{N}\sum_{i}I(|d_i-d_i^{*}|<x)$\\
\bottomrule
\end{tabular}
}
\caption{Quantitative metrics for depth estimation. $d_i$ is the predicted depth; $d_i^*$ is the ground truth depth; $N$ corresponds to all pixels with the ground-truth label. $I$ is the indicator function.}
\label{table:metric}
\end{table}

\begin{table*}[!t]
\centering
\resizebox{0.95\textwidth}{!}{%
\begin{tabular}{r|ccccc|ccc}
\toprule
\multicolumn{1}{c|}{Method}       & AbsRel $\downarrow$ & SqRel $\downarrow$& log10 $\downarrow$& RMSE $\downarrow$  & RMSELog $\downarrow$& $\delta<1.25$ $\uparrow$& $\delta<1.25^2$ $\uparrow$& $\delta<1.25^3$ $\uparrow$ \\ \hline
Bts~\cite{lee2019big}    & 0.117 &0.052& 0.049 & 0.270 & 0.151 & 0.862 & 0.966 & 0.992\\ 
Bts$^*$~\cite{lee2019big}    & 0.088 &0.035& 0.038 & 0.228 & 0.128 & 0.916 & 0.980 & 0.994\\ \hline
MVSNet~\cite{yao2018mvsnet}  & 0.094 & 0.042  &   0.040  &   0.251  &   0.135  &   0.897  &   0.975  &0.993\\ 
FastMVS~\cite{yu2020fast} & 0.089  &   0.038  &   0.038  &   0.231  &   0.128  &   0.912  &   0.978 & 0.993 \\
DPSNet~\cite{im2019dpsnet}  & 0.094  &   0.041  &   0.043  &   0.258  &   0.141  &   0.883  &   0.970 & 0.992\\
NAS~\cite{kusupati2019normal}  
& 0.086  &   0.032  &   0.038  &   0.224  &   0.122  &   0.917  &   0.984 & 0.996\\ 
PatchmatchNet~\cite{wang2021patchmatchnet} &  0.133  &    0.075  &    0.055  &    0.320  &    0.175  &    0.834  &    0.955  &    0.987 \\
\hline
Ours-mono  & 0.145 & 0.065  & 0.061  &   0.300   &  0.173   &  0.807 & 0.957 & 0.990\\ 
Ours-mono$^*$  & 0.103 & 0.037  & 0.044  &   0.237   &  0.135   &  0.892 & 0.984 & 0.996\\ 
Ours-robust &    \textbf{0.059}  &    \textbf{0.016}  &    \textbf{0.026}  &    \textbf{0.159}  &    \textbf{0.083}  &    \textbf{0.965}  &    \textbf{0.996}  &    \textbf{0.999}  \\
Ours&    \textbf{0.059}  &    0.017  &    \textbf{0.026}  &    0.162  &    0.084  &    0.963  &    0.995  &    \textbf{0.999}  \\ 
\bottomrule
\end{tabular}
}
\vspace{-0.1in}
\caption{Depth evaluation results on ScanNet~\cite{scannet17cvpr}. We compare against both multi-view depth estimation methods~\cite{yu2020fast, yao2018mvsnet, im2019dpsnet,kusupati2019normal,wang2021patchmatchnet} and a state-of-the-art single-view method~\cite{lee2019big}. Our approach achieve significant improvements over top-performing method NAS~\cite{kusupati2019normal} on AbsRel. The improvements are consistent across all metrics.}
\label{table:scannet}
\vspace{-0.1in}
\end{table*}

\begin{table*}
\centering
\resizebox{0.95\textwidth}{!}{%
\begin{tabular}{cr|ccccc|ccc}
\toprule
& \multicolumn{1}{c|}{Method}       & AbsRel $\downarrow$& AbsDiff $\downarrow$& SqRel $\downarrow$& RMSE  $\downarrow$& RMSELog $\downarrow$& $\delta<1.25$ $\uparrow$& $\delta<1.25^2$ $\uparrow$& $\delta<1.25^3$ $\uparrow$\\ \hline

\parbox[t]{2mm}{\multirow{6}{*}{\rotatebox[origin=c]{90}{SUN3D (Real)}}}  
& COLMAP~\cite{sfm16cvpr}  &  0.623 & 1.327 & 3.236 & 2.316 & 0.661 & 0.327 & 0.554 & 0.718\\
& DeMoN~\cite{ummenhofer17demon}   &  0.214 & 2.148 & 1.120 & 2.421 & 0.206 & 0.733 & 0.922 & 0.963\\
& DeepMVS~\cite{huang18deepmvs} &  0.282 & 0.604 & 0.435 & 0.944 & 0.363 & 0.562 & 0.739 & 0.895\\ 
& DPSNet-U~\cite{im2019dpsnet} &  0.147 & 0.336 & 0.117 & 0.449 & 0.196 & 0.781 & 0.926 & 0.973\\
& NAS~\cite{kusupati2019normal}    &   0.127 & 0.288 & 0.085 & 0.378 & 0.170 & 0.830 & 0.944 & 0.978\\
& Ours-robust & \underline{0.100}  &  \underline{0.231}  & \underline{0.057}   &   \underline{0.313}  &   \underline{0.140}  &   \textbf{0.895}  &   \underline{0.966}  &   \underline{0.991}  \\
& Ours &\textbf{0.099}& \textbf{0.224} &   \textbf{0.055}  &   \textbf{0.304}  &   \textbf{0.137}  &   \underline{0.893}  &   \textbf{0.970}  &   \textbf{0.993} \\

\hline
\parbox[t]{2mm}{\multirow{6}{*}{\rotatebox[origin=c]{90}{RGBD (Real)}}} 
& COLMAP~\cite{sfm16cvpr}  & 0.539 & 0.940 & 1.761 & 1.505 & 0.715 & 0.275 & 0.500 & 0.724\\
& DeMoN~\cite{ummenhofer17demon}  &  0.157 & 1.353 & 0.524 & 1.780 & 0.202 & 0.801 & 0.906 & \textbf{0.962}\\
    & DeepMVS~\cite{huang18deepmvs} &  0.294 & 0.621 & 0.430 & 0.869 & 0.351 & 0.549 & 0.805 & 0.922\\
    & DPSNet-U~\cite{im2019dpsnet} &  0.151 & 0.531 & 0.251 & 0.695 & 0.242 & 0.804 & 0.895 & 0.927\\
    & NAS~\cite{kusupati2019normal} &   0.131 & 0.474 & 0.213 & 0.619 & 0.209 & 0.857 & 0.929 & 0.945\\
& Ours-robust &   \textbf{0.078}  &   \textbf{0.311} & \textbf{0.156}  &   \underline{0.443}  &   \textbf{0.146}  &   \textbf{0.926}  &   \textbf{0.945}  &   \underline{0.954} \\
& Ours &   \underline{0.082}  &    \underline{0.325} & \underline{0.165}  &   \textbf{0.440}  &   \underline{0.147}  &   \underline{0.921}  &   \underline{0.939}  &   0.948 \\ 
\hline
\parbox[t]{2mm}{\multirow{6}{*}{\rotatebox[origin=c]{90}{Scenes11 (Syn)}}} & COLMAP~\cite{sfm16cvpr}  &  0.625 & 2.241 & 3.715 & 3.658 & 0.868 & 0.390 & 0.567 & 0.672\\
& DeMoN~\cite{ummenhofer17demon}   &  0.556 & 1.988 & 3.402 & 2.603 & 0.391 & 0.496 & 0.726 & 0.826\\
& DeepMVS~\cite{huang18deepmvs} &  0.210 & 0.597 & 0.373 & 0.891 & 0.270 & 0.688 & 0.894 & 0.969\\
& DPSNet~\cite{im2019dpsnet} &  0.050 & 0.152 & 0.111 & 0.466 & 0.116 & 0.961 & 0.982 & 0.988\\
&  NAS~\cite{kusupati2019normal}  &  \textbf{0.038} & \textbf{0.113} & \underline{0.067} & \textbf{0.371} & \textbf{0.095} & \underline{0.975} & \underline{0.990} & \textbf{0.995}\\ 
 & Ours-robust & \underline{0.041}  & \underline{0.141} &  \textbf{0.066}   &   \underline{0.410}  &   \underline{0.099}  &   \textbf{0.979}  &   \textbf{0.991}  &   \underline{0.994}  \\
& Ours &   0.046  &   0.155 & 0.080     &   0.439  &   0.107  &   0.976  &   0.989  &   0.993  \\
\bottomrule                                         
\end{tabular}
}
\vspace{-0.1in}
\caption{Depth evaluation results on SUN3D, RGBD, and Scenes11 datasets(synthetic). The numbers for COLMAP, DeMoN, DeepMVS, DPSNet, and NAS are obtained from \cite{kusupati2019normal}. We achieve significant improvements on SUN3D and RGBD. We show the best number in bold and the second best with underline.}
\label{table:sun3d}
\vspace{-0.1in}
\end{table*}

\subsection{Baseline Approaches}\label{sec4.3} 

\myparagraph{MVSNet~\cite{yao2018mvsnet}} is an end-to-end plane sweeping stereo approach based on 3D-cost volume. 

\myparagraph{DPSNet~\cite{im2019dpsnet}} shares similar spirit of MVSNet~\cite{yao2018mvsnet} but focus on accurate depth map prediction.

\myparagraph{NAS~\cite{kusupati2019normal}} is a recent work that jointly predicts consistent depth and normal, using extra normal supervision.

\myparagraph{FastMVSNet~\cite{yu2020fast}} is a recent variant to MVSNet which accelerate the computation by computing sparse cost volume.

\myparagraph{Bts~\cite{lee2019big}} is a state-of-the-art single view depth prediction network. It incorporates planar priors into network design. 
Additionally, we use an asterisk sign `$*$' to denote an oracle version \textbf{Bts$^*$}, where we use the ground truth depth map to factor out the global scale. 

\myparagraph{PatchmatchNet~\cite{wang2021patchmatchnet}} is one of the most recent state-of-the-art efficient MVS algorithm.

\myparagraph{Ours-mono} is our method without the epipolar attention module, thus equivalent to single-view depth estimation. Similar to Bts$^*$, we also report the results factoring out the global scale for \textbf{Ours-mono$^*$}.

\myparagraph{Ours-robust} is our method with multi-scale epipolar attention module applied on $\mathcal{F}$. 

\myparagraph{Ours} is our method with epipolar attention module applied only in $\mathcal{F}$'s second layer.

\begin{figure*}[!t]
\newcommand{\TT}[1]{\raisebox{-0.5\height}{#1}}
\setlength{\tabcolsep}{1pt}
\footnotesize
\def\imw{0.16\textwidth}
\begin{tabular}{cccccc}
Bts~\cite{lee2019big} & FastMVSNet~\cite{yu2020fast} & 
NAS~\cite{kusupati2019normal} & PatchmatchNet~\cite{wang2021patchmatchnet} & \ouralg (Ours) & G.T. \\
\TT{\includegraphics[width=\imw]{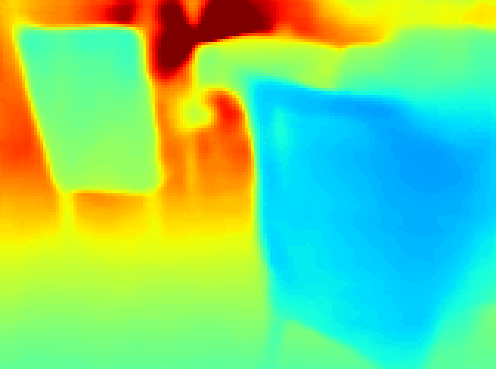}} & 
\TT{\includegraphics[width=\imw]{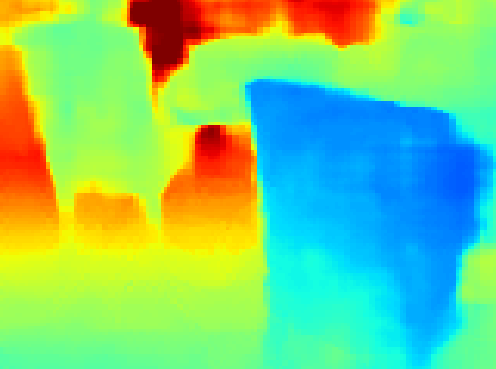}} & 
\TT{\includegraphics[width=\imw]{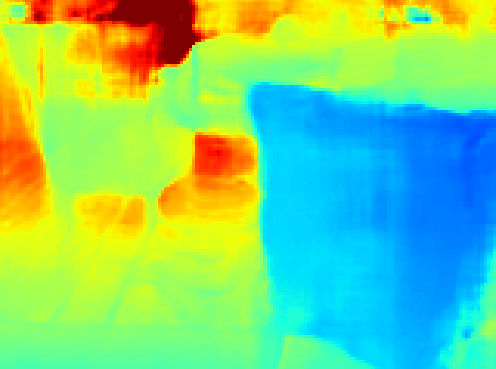}} & 
\TT{\includegraphics[width=\imw]{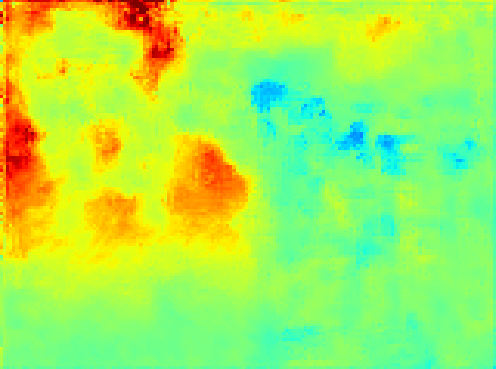}} & 
\TT{\includegraphics[width=\imw]{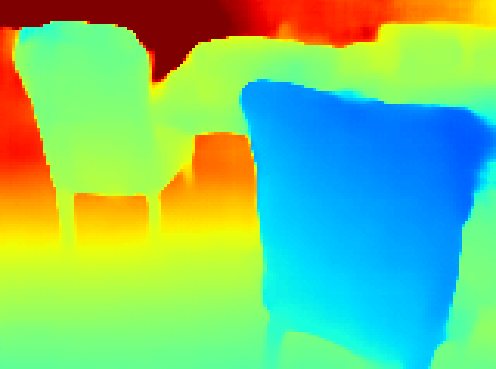}} & 
\TT{\includegraphics[width=\imw]{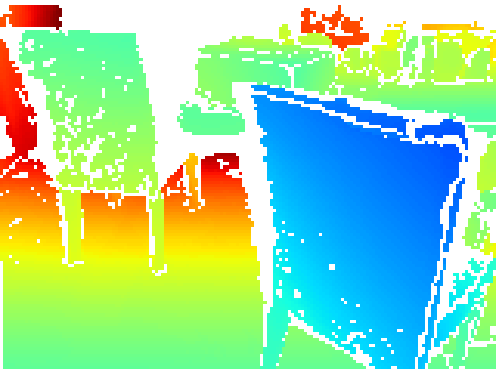}} \\

\TT{\includegraphics[width=\imw]{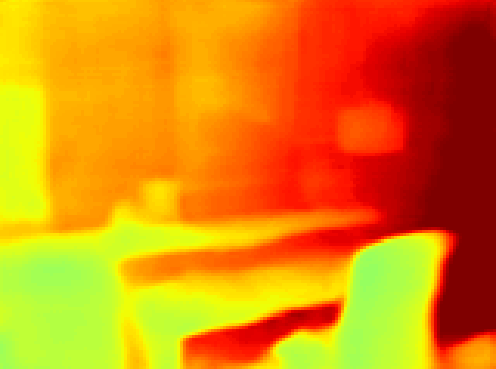}} & 
\TT{\includegraphics[width=\imw]{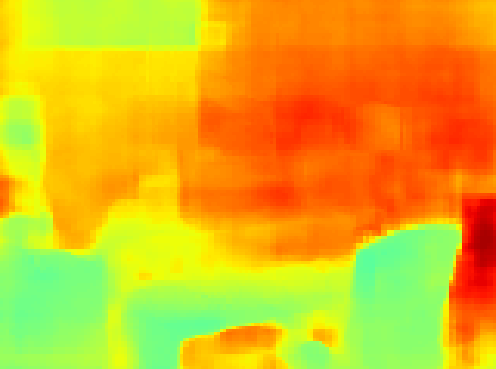}} & 
\TT{\includegraphics[width=\imw]{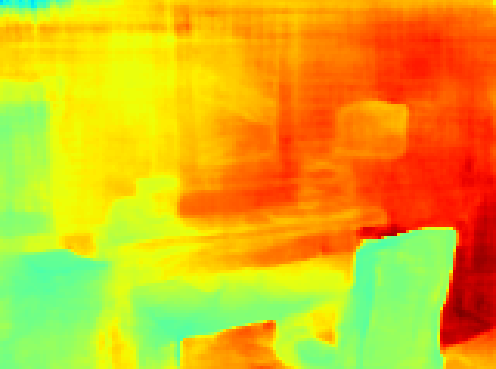}} & 
\TT{\includegraphics[width=\imw]{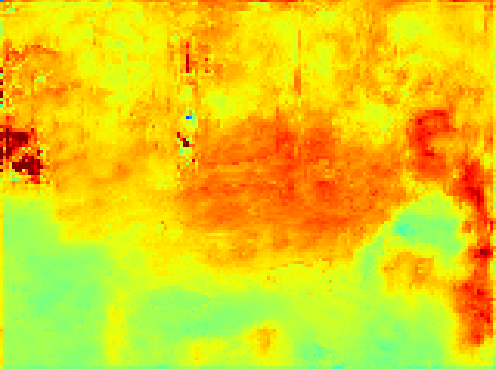}} & 
\TT{\includegraphics[width=\imw]{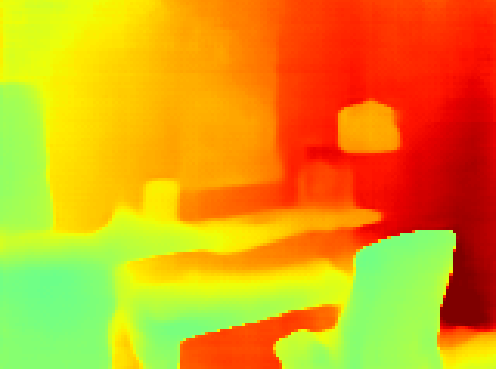}} & 
\TT{\includegraphics[width=\imw]{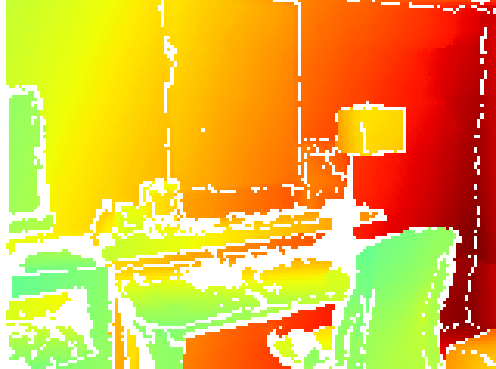}} \\

\TT{\includegraphics[width=\imw]{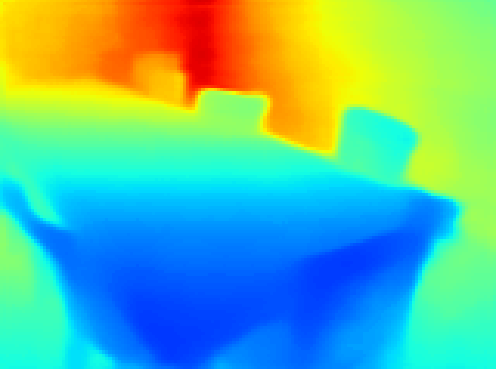}} & 
\TT{\includegraphics[width=\imw]{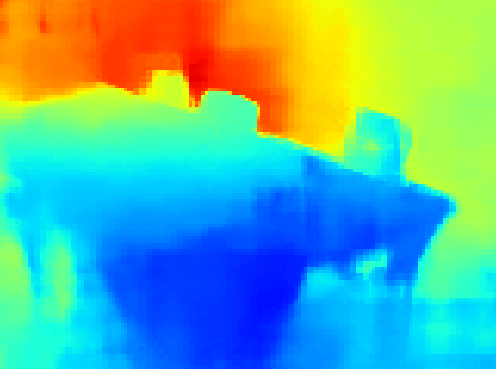}} & 
\TT{\includegraphics[width=\imw]{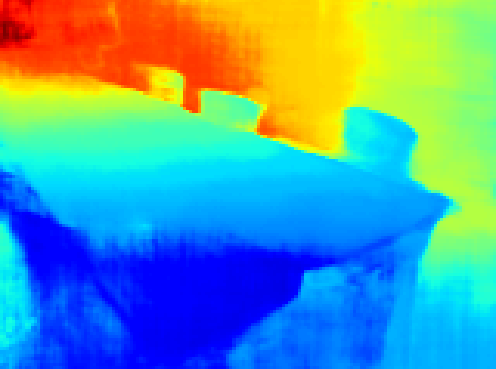}} & 
\TT{\includegraphics[width=\imw]{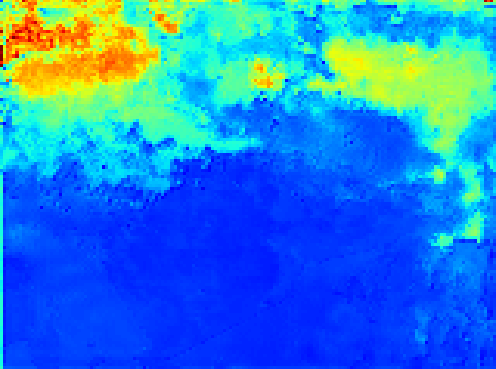}} & 
\TT{\includegraphics[width=\imw]{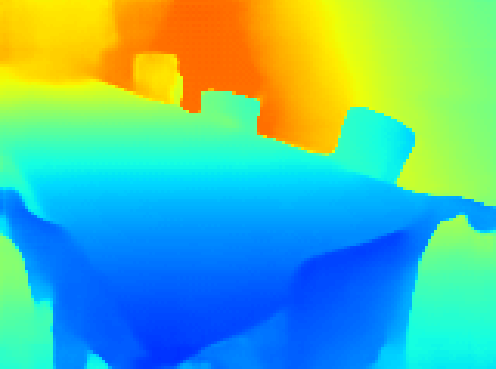}} & 
\TT{\includegraphics[width=\imw]{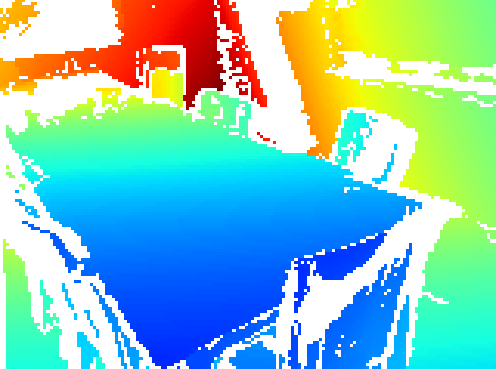}} \\
\end{tabular}
\caption{Qualitative results on depth prediction. Each row corresponds to one test example. The region without ground truth depth labels is colored white in GT. Our prediction outperforms both the single-view depth estimation method~\cite{lee2019big} and other multi-view methods. }
\label{fig:qualitative}
\vspace{-0.1in}
\end{figure*}

\begin{figure*}
\begin{center}
\def\imw{0.3\textwidth}
\begin{tabular}{ccc}
\subfigimg[width=\imw]{PatchmatchNet}{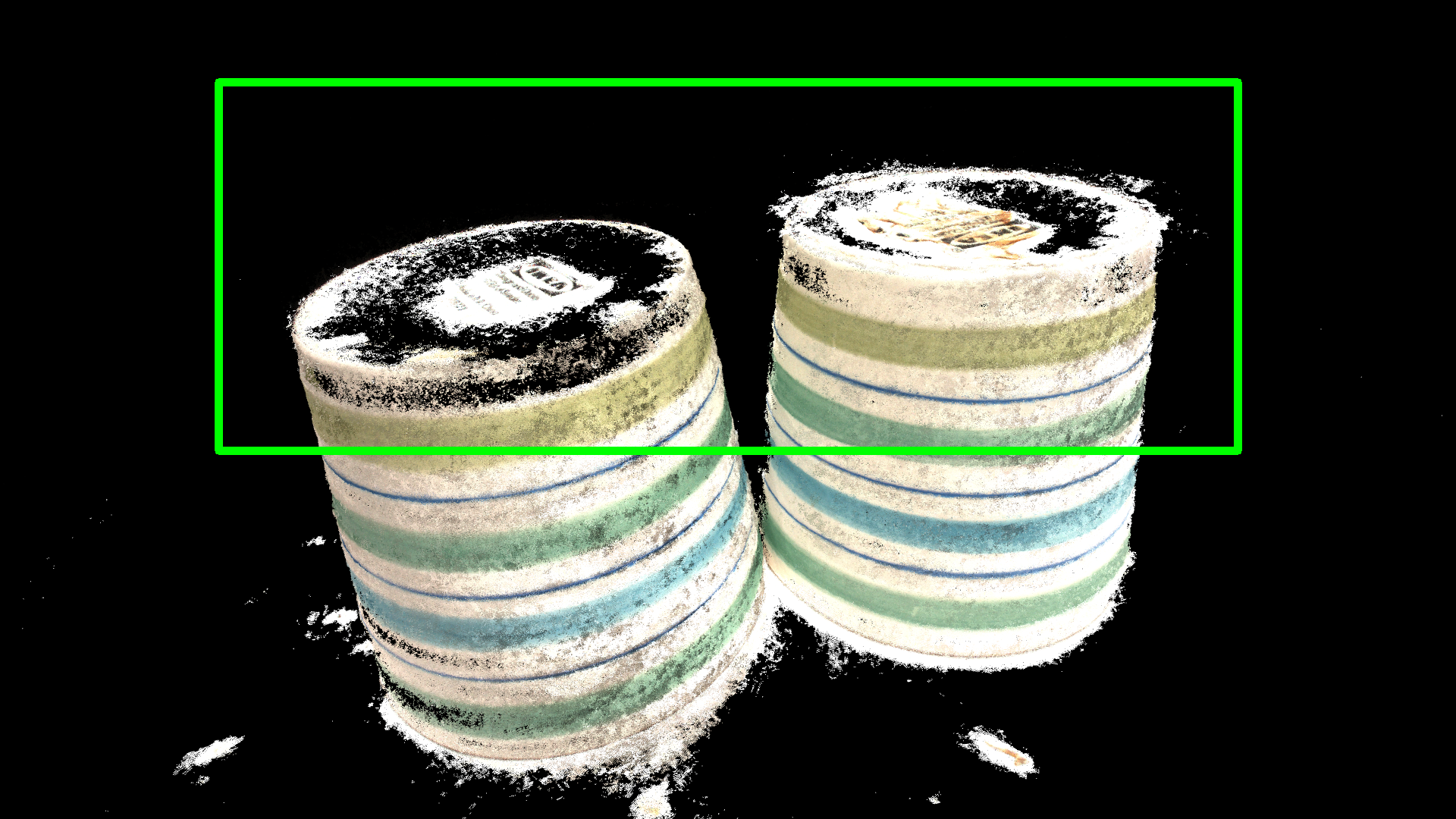}&\subfigimg[width=\imw]{CasMVSNet}{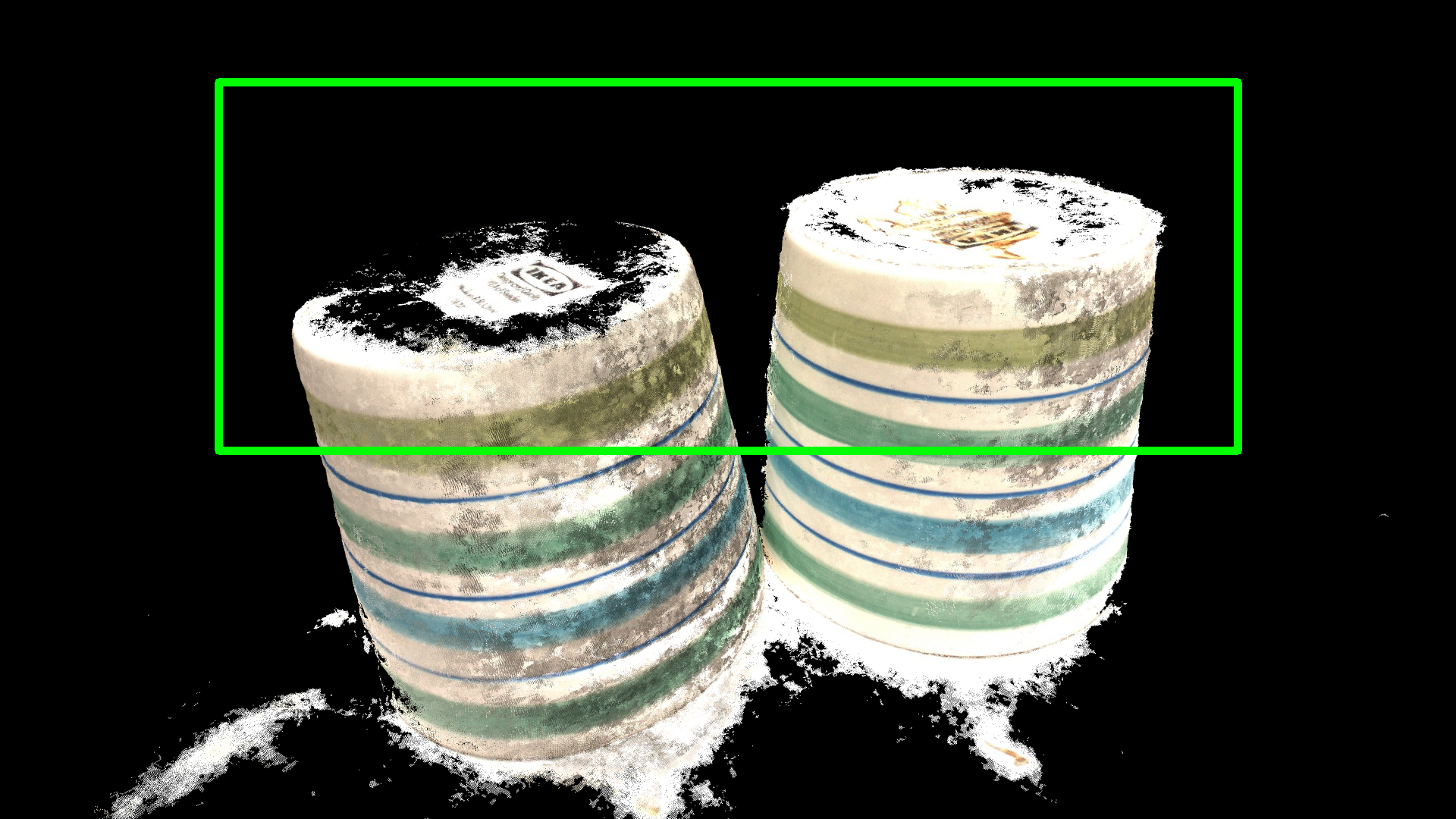}&\subfigimg[width=\imw]{MVS2D (ours)}{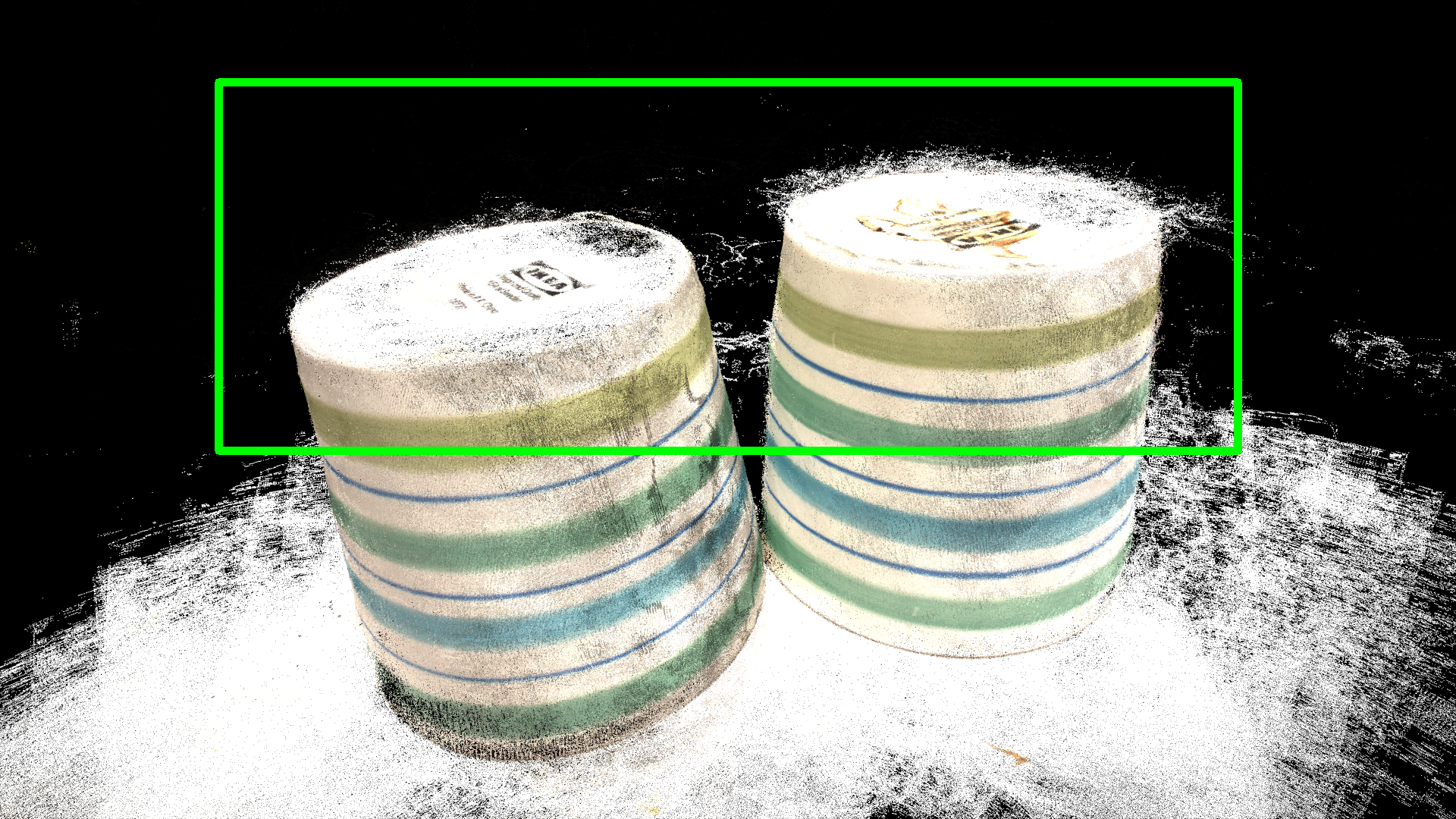}\\
\subfigimg[width=\imw]{PatchmatchNet}{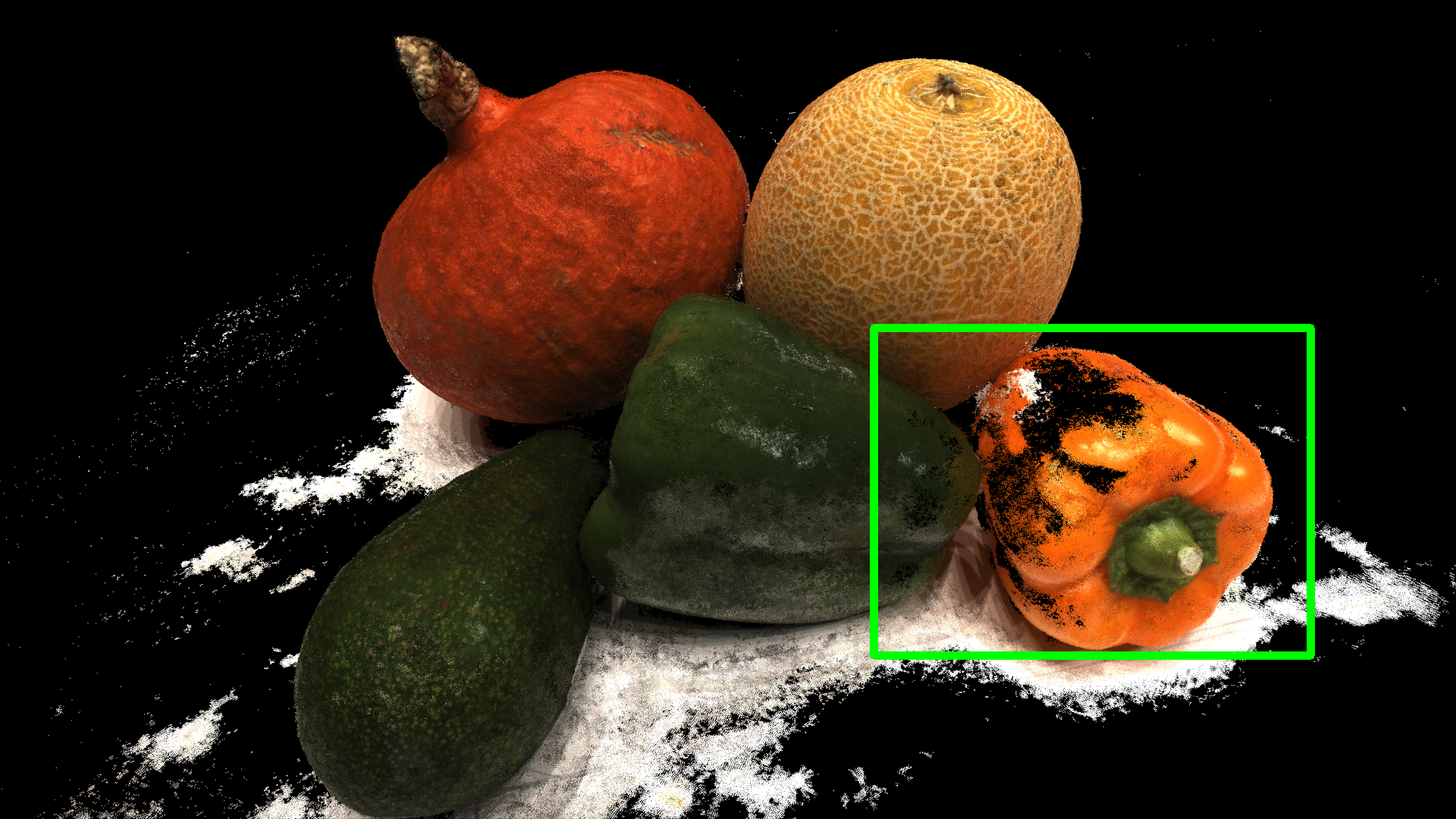}&\subfigimg[width=\imw]{CasMVSNet}{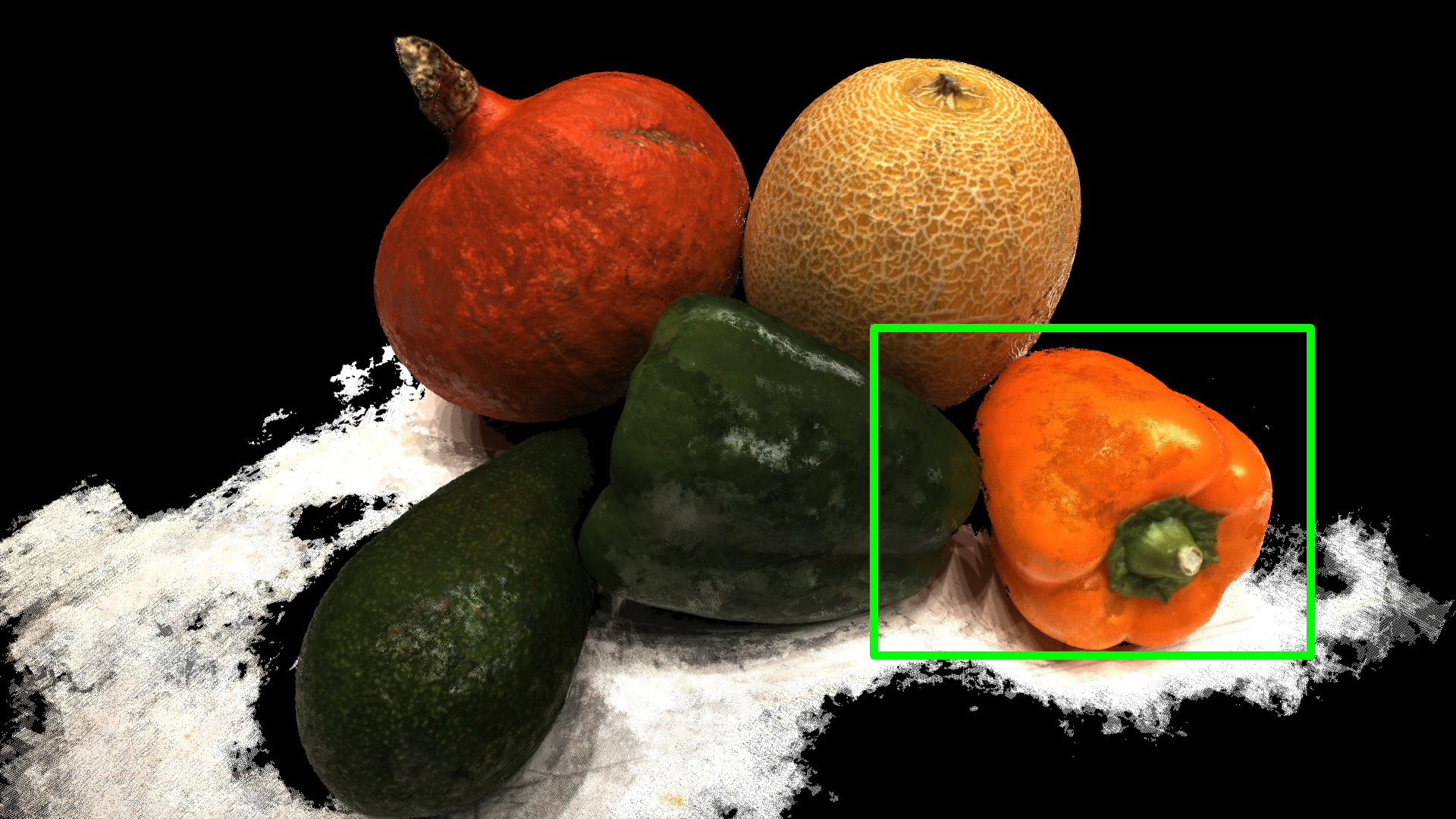}&\subfigimg[width=\imw]{MVS2D (ours)}{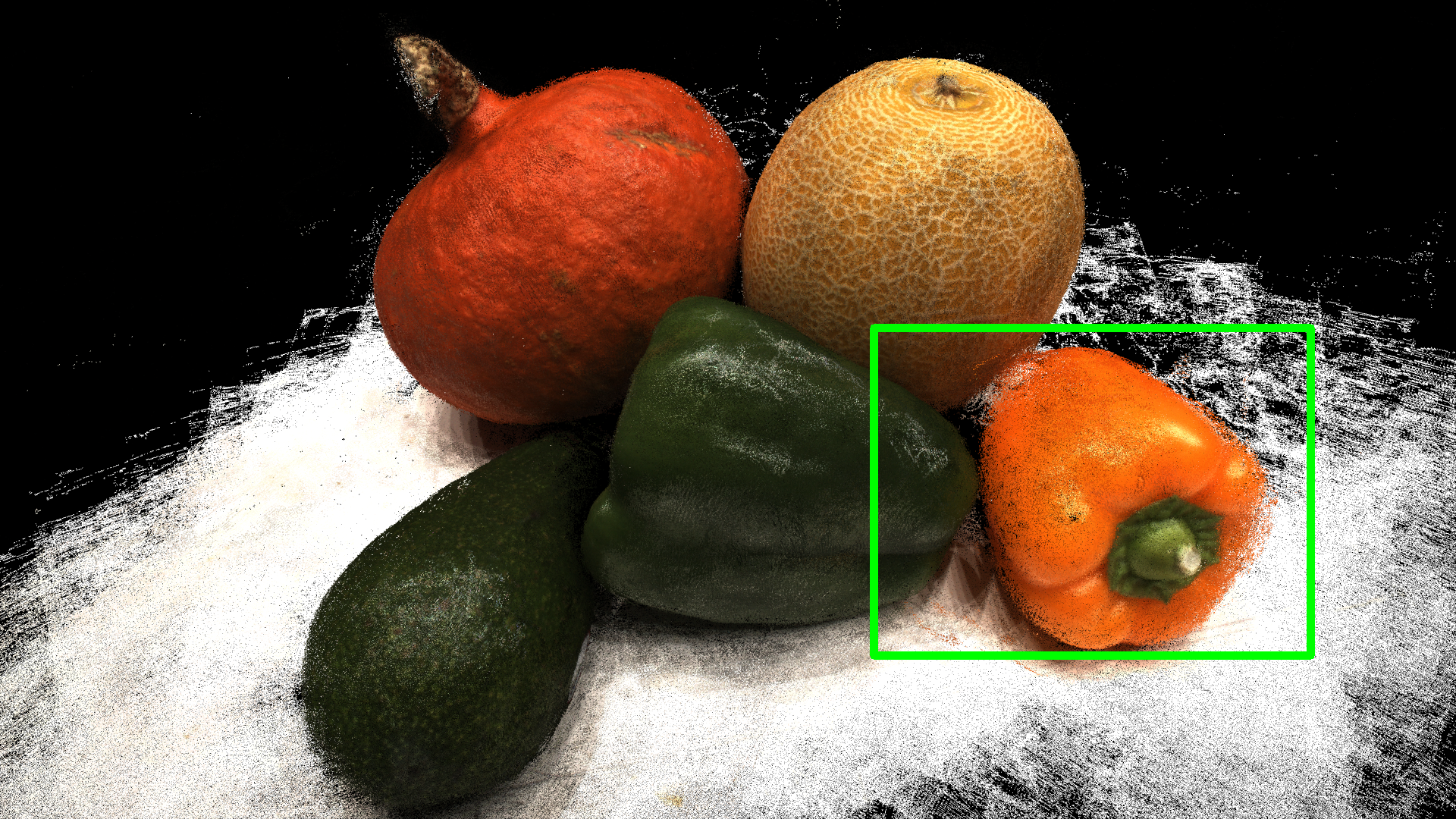}
\end{tabular}
\vspace{-0.1in}
\caption{Qualitative 3D reconstruction results on DTU dataset. \ouralg produces more complete reconstruction in texture-less region.}
\vspace{-0.2in}
\label{fig:dtu_3d}
\end{center}
\end{figure*}

\subsection{Result Analysis}\label{sec4.4}
\myparagraph{Comparison on Efficiency.} 
We compare against both single-view methods~\cite{lee2019big} and multi-view methods~\cite{yao2018mvsnet,yu2020fast,im2019dpsnet,kusupati2019normal}. The inference speed of our method is comparable to single-view methods~\cite{lee2019big} and significantly outperforms other multi-view methods~\cite{yao2018mvsnet,yu2020fast,im2019dpsnet,kusupati2019normal}. Evaluations are done on ScanNet~\cite{scannet17cvpr}. Our method is $48\times$ faster than NAS, $39\times$ faster than DPSNet, $10\times$ faster than MVSNet, and $4\times$ faster than the FastMVSNet. Please refer to the supp. for more details.

\myparagraph{Comparison on Depth Estimation.}
\ouralg achieves considerable improvements in the depth prediction accuracy (see Table~\ref{table:scannet}). On ScanNet, our approach outperforms MVSNet by large margins, reducing AbsRel error from \textbf{0.094} to \textbf{0.059}. The improvements are consistent across most other metrics. Remarkably, our approach also outperforms NAS, which uses more parameters and runs 48 times slower. We visualize some depth predictions in Figure~\ref{fig:qualitative}.

Our approach yields significant improvements over single-view baselines. Adding multi-view cues improves the AbsRel of ours-mono from \textbf{0.145} to \textbf{0.059} on ScanNet. Since single-view has scale-ambiguity, we further investigate whether our methods will still be favorable when factoring out the scale. The results show that when eliminating the scale, ours-mono$^*$ has AbsRel 0.103, which is still a significant improvement. This means our approach does not simply infer the global scaling factor from multi-view cues. Compared with the single-view model, our model only incurs a $5.8\% $ increase in parameters. Such efficiency will enable multi-view methods to embrace a much larger 2D convolutional network which is not possible before. 

On other datasets, \ouralg also performs favorably (see Table~\ref{table:sun3d}). We achieve the AbsRel error of 0.078 on the RGBD dataset, while the next best NAS only achieves 0.131. Although \ouralg excels at adapting to the scene prior, it is encouraging that it also performs well on the Scenes11 dataset, a synthetic scene with randomly placed objects. We ranked second on the Scenes11 dataset on AbsRel. Please refer to the supp. material for experiments on our generalization ability to novel datasets.

\myparagraph{Evaluations on DTU.} We evaluate on DTU dataset following the practice of \cite{wang2021patchmatchnet}. We use 4 reference views and 96 depth samples uniformly placed in the inverse depth space ([$\frac{1}{935.}$,$\frac{1}{425.}$]). 
The quantitative results can be found in Table \ref{tab:evaluation_dtu}. \ouralg is the best on overall score and the second-best completeness score. Such performance is encouraging since our method is quite simple: it is just a single-stage procedure without using any multi-stage refinement as commonly used in recent MVS algorithms (\cite{yao2019recurrent,wang2021patchmatchnet, gu2020cascade}). We show some qualitative results of 3D reconstruction on DTU objects in Figure \ref{fig:dtu_3d}. Qualitatively, our reconstruction is typically more complete on flat surface areas. The behavior is reasonable because our approach utilizes strong single-view priors. We also compare the inference speed with the recent SOTA PatchmatchNet\cite{wang2021patchmatchnet}. Our approach yield around 2x speed up as shown in Table \ref{table:dtu-fps}. Lastly, as our method was mainly designed for multi-view depth estimation, we additionally examine the depth evaluation metrics. Since DTU does not have ground truth depth for the test set, we report the depth evaluation results on the validation set. As expected, \ouralg is better than PatchmatchNet in terms of depth metrics, and the performance gap there is wider than in 3D Reconstruction. The results can be found in Table \ref{table:dtu-depth}.

\begin{table}[t]
\setlength{\abovecaptionskip}{0.1cm}
\centering
\footnotesize
\begin{widetable}{\columnwidth}{cccc}
 \toprule
 Methods & Acc.(mm) & Comp.(mm) & Overall(mm)\\
 \hline
 Camp~\cite{campbell2008using} & 0.835 & 0.554 & 0.695 \\ 
 Furu~\cite{furukawa2009accurate} & 0.613 & 0.941 & 0.777 \\
 Tola~\cite{tola2012efficient} & 0.342 & 1.190 & 0.766 \\
 Gipuma~\cite{galliani2015massively} & \textbf{0.283} & 0.873 & 0.578 \\
 SurfaceNet~\cite{ji2017surfacenet}  & 0.450 & 1.040 & 0.745 \\
 MVSNet~\cite{yao2018mvsnet} & 0.396 & 0.527 & 0.462\\
 R-MVSNet~\cite{yao2019recurrent} & 0.383 & 0.452 & 0.417\\
 CIDER~\cite{xu2020learning} & 0.417 & 0.437 & 0.427\\
 P-MVSNet~\cite{luo2019p} & 0.406 & 0.434 & 0.420\\
 Point-MVSNet~\cite{chen2019point} & 0.342 & 0.411 & 0.376\\
 Fast-MVSNet~\cite{yu2020fast} & 0.336 & 0.403 & 0.370\\ 
 CasMVSNet~\cite{gu2020cascade} & 0.325 & 0.385 & 0.355\\
 UCS-Net~\cite{cheng2020deep} & 0.338 & 0.349 & \underline{0.344} \\
 CVP-MVSNet~\cite{yang2020cost} & \underline{0.296} & 0.406 & 0.351 \\
 PatchMatchNet~\cite{wang2021patchmatchnet} & 0.427 & \textbf{0.277} & 0.352\\
  \ouralg(Ours) & 0.394 & \underline{0.290} & \textbf{0.342}\\
 \bottomrule
\end{widetable}
\caption{Quantitative results on the evaluation set of DTU~\cite{aanaes2016large}. We bold the best number and underline the second best number.}
\label{tab:evaluation_dtu}
\end{table}

\begin{table}
\centering
\resizebox{0.95\linewidth}{!}{%
\begin{tabular}{rccc}
\toprule
\multicolumn{1}{c}{Metric} & FPS$640\times 480$ $\uparrow$ &FPS$1280\times 640$ $\uparrow$ & FPS$1536\times 1152$ $\uparrow$     \\ \hline
\multicolumn{1}{c}{PatchmatchNet }  &16.5  &  6.30 & 4.57 \\  
\multicolumn{1}{c}{\ouralg(Ours) } &\textbf{36.4}  &  \textbf{10.9} &  \textbf{7.3}   \\  
\bottomrule
\end{tabular}
}
\vspace{-0.1in}
\caption{Speed benchmark on DTU dataset. We show FPS (frame per second) on three input resolutions. We use one source image and 4 reference images. }
\vspace{-0.1in}
\label{table:dtu-fps}
\end{table}

\begin{table}
\centering
\resizebox{0.95\linewidth}{!}{%
\begin{tabular}{rcccc}
\toprule
\multicolumn{1}{c}{Metric} & RMSE(mm)$\downarrow$ & thre@0.2$\uparrow$ & thre@0.5$\uparrow$ & thre@1.0$\uparrow$      \\ \hline
\multicolumn{1}{c}{PatchmatchNet } & 32.348 &0.169  &    0.387  &    0.610\\  
\multicolumn{1}{c}{\ouralg(Ours) } & \textbf{14.769} & \textbf{0.238}  &    \textbf{0.504}  &    \textbf{0.718} \\  
\bottomrule
\end{tabular}
}
\vspace{-0.1in}
\caption{Depth evaluation on DTU validation set. We show the root mean square error and the percentage of errors fall below 0.2/0.5/1.0mm thresholds.}
\vspace{-0.2in}
\label{table:dtu-depth}
\end{table}

\myparagraph{Comparison on Robustness under Noisy Pose.} As shown in Table \ref{table:scannet}, Ours-robust (multi-scale cues) and Ours (single-scale cues) perform similarly when the input poses are accurate. However, as shown in Table \ref{table:perturb}, multi-scale aggregation is preferred when the input poses are noisy. It suggests that when having inaccurate training data, it is necessary to incorporate multi-scale cues, though at a cost of increased computations (as shown in Table~\ref{table:speed}).

\begin{table}[!h]
\centering
\resizebox{0.95\linewidth}{!}{%
\begin{tabular}{rcccc}
\toprule
\multicolumn{1}{c}{Metric} & MVSNet& DPSNet & Ours & Ours-robust      \\ \hline
\multicolumn{1}{c}{AbsRel $\downarrow$} & 0.094 &0.094 & 0.059 & \textbf{0.059}\\  
\multicolumn{1}{c}{AbsRel (p) $\downarrow$} & 0.113 &0.126 & 0.073  & \textbf{0.070}  \\  
\multicolumn{1}{c}{$\Delta$ $\downarrow$} & 0.019 &0.032  & 0.014    & \textbf{0.011} \\ \hline 
\multicolumn{1}{c}{$\delta<1.25$ $\uparrow$} & 0.897 & 0.871 & 0.983 & \textbf{0.965}  \\ 
\multicolumn{1}{c}{$\delta<1.25$ (p) $\uparrow$} & 0.851 & 0.807  & 0.947  & \textbf{0.952}\\ 
\multicolumn{1}{c}{$\Delta$ $\downarrow$} & 0.046 & 0.064 & 0.016   & \textbf{0.013}  \\
\bottomrule
\end{tabular}
}
\caption{Different methods' performance under noisy input poses on ScanNet~\cite{scannet17cvpr}. We notice that most methods suffer from significant performance drops. Our method with multi-scale epipolar aggregation shows notable robustness.}
\label{table:perturb}
\end{table}

\myparagraph{Ablation Study on Depth Encoding.} 
The ablation study of our depth code design can be found in Table \ref{table:embed}. We tested four code types. `Uniform' serves as a sanity check, where we use the same code vector for all depth hypotheses. In other words, the network does not extract useful information from the reference images. `Linear' improves on uniform encoding by scaling a base code vector with the corresponding depth value. `Cosine' codes are identical to the one used in \cite{vaswani2017attention}. `Learned` codes are optimized end-to-end. We can see that learning the codes end-to-end leads to noticeable performance gains. One explanation is that these learned codes can adapt to the single-view feature representations of the source image.

\begin{table}[!h]
\centering
\resizebox{0.95\linewidth}{!}{%
\begin{tabular}{rcccc}
\toprule
\multicolumn{1}{c}{Metric} & Uniform & Linear & Cosine  & Learned    \\ \hline
\multicolumn{1}{c}{AbsRel $\downarrow$} & 0.139 & 0.128& 0.064 & \textbf{0.059}      \\ 
\multicolumn{1}{c}{$\delta < 1.25$ $\uparrow$} &  0.815 & 0.840  & 0.961 & \textbf{0.964}   \\ 
\multicolumn{1}{c}{RMSE $\downarrow$} &  0.293 &0.283   &  0.166& \textbf{0.156 }    \\ 

\bottomrule
\end{tabular}
}
\caption{Ablation study on different depth encodings. We can see that jointly training depth encodings gives the best performance. } 
\label{table:embed}
\vspace{-0.1in}
\end{table}

%% file: 05_conclusions.tex
\section{Conclusions and Limitations}

\noindent\textbf{Conclusions.} We proposed a simple yet effective method for multi-view stereo. The core of our method is to integrate single-view and multi-view cues during the prediction jointly. Such a design not only improves the performance but also has the appealing factor of being efficient. Furthermore, we have demonstrated the trade-off between input pose accuracy and network complexity. When the input pose is exact, we can leverage minimum additional computation to inject more multi-view information through the epipolar attention. 

\noindent\textbf{Limitations.} One limitation of our approach is that the network is trained in a way that adapted to data distribution well, which might makes it less generalizable to out-of-distribution testing data. In the future, we propose to address this issue by developing robust training losses. Another limitation is that the proposed attention mechanism does not explicitly model the consistency between different pixels on the same epipolar line. We plan to address this issue by developing novel attention mechanisms to explicitly enforce those constraints. 

%% file: main.bbl
\begin{thebibliography}{10}\itemsep=-1pt

\bibitem{aanaes2016large}
Henrik Aan{\ae}s, Rasmus~Ramsb{\o}l Jensen, George Vogiatzis, Engin Tola, and
  Anders~Bjorholm Dahl.
\newblock Large-scale data for multiple-view stereopsis.
\newblock {\em International Journal of Computer Vision (IJCV)},
  120(2):153--168, 2016.

\bibitem{badki2020bi3d}
Abhishek Badki, Alejandro Troccoli, Kihwan Kim, Jan Kautz, Pradeep Sen, and
  Orazio Gallo.
\newblock {Bi3D}: Stereo depth estimation via binary classifications.
\newblock In {\em Proceedings of the IEEE Conference on Computer Vision and
  Pattern Recognition (CVPR)}, pages 1600--1608, 2020.

\bibitem{campbell2008using}
Neill~DF Campbell, George Vogiatzis, Carlos Hern{\'a}ndez, and Roberto Cipolla.
\newblock Using multiple hypotheses to improve depth-maps for multi-view
  stereo.
\newblock In {\em Proceedings of the European Conference on Computer Vision
  (ECCV)}, pages 766--779. Springer, 2008.

\bibitem{chen2019point}
Rui Chen, Songfang Han, Jing Xu, and Hao Su.
\newblock Point-based multi-view stereo network.
\newblock In {\em Proceedings of the IEEE International Conference on Computer
  Vision (ICCV)}, pages 1538--1547, 2019.

\bibitem{cheng2018geometry}
Ricson Cheng, Ziyan Wang, and Katerina Fragkiadaki.
\newblock Geometry-aware recurrent neural networks for active visual
  recognition.
\newblock In {\em Advances in Neural Information Processing Systems (NeurIPS)},
  pages 5086--5096, 2018.

\bibitem{cheng2020deep}
Shuo Cheng, Zexiang Xu, Shilin Zhu, Zhuwen Li, Li~Erran Li, Ravi Ramamoorthi,
  and Hao Su.
\newblock Deep stereo using adaptive thin volume representation with
  uncertainty awareness.
\newblock In {\em Proceedings of the IEEE Conference on Computer Vision and
  Pattern Recognition (CVPR)}, pages 2524--2534, 2020.

\bibitem{scannet17cvpr}
Angela Dai, Angel~X. Chang, Manolis Savva, Maciej Halber, Thomas Funkhouser,
  and Matthias Nie{\ss}ner.
\newblock {ScanNet}: Richly-annotated {3D} reconstructions of indoor scenes.
\newblock In {\em Proceedings of the IEEE Conference on Computer Vision and
  Pattern Recognition (CVPR)}, 2017.

\bibitem{duggal2019deeppruner}
Shivam Duggal, Shenlong Wang, Wei-Chiu Ma, Rui Hu, and Raquel Urtasun.
\newblock {DeepPruner}: Learning efficient stereo matching via differentiable
  patchmatch.
\newblock In {\em Proceedings of the IEEE International Conference on Computer
  Vision (ICCV)}, pages 4384--4393, 2019.

\bibitem{Furukawa:2015:MST}
Yasutaka Furukawa and Carlos Hern\'{a}ndez.
\newblock Multi-view stereo: A tutorial.
\newblock {\em Found. Trends. Comput. Graph. Vis.}, 9(1-2):1--148, June 2015.

\bibitem{furukawa2009accurate}
Yasutaka Furukawa and Jean Ponce.
\newblock Accurate, dense, and robust multiview stereopsis.
\newblock {\em IEEE Transactions on Pattern Analysis and Machine Intelligence
  (TPAMI)}, 32(8):1362--1376, 2009.

\bibitem{galliani2015massively}
Silvano Galliani, Katrin Lasinger, and Konrad Schindler.
\newblock Massively parallel multiview stereopsis by surface normal diffusion.
\newblock In {\em Proceedings of the IEEE International Conference on Computer
  Vision (ICCV)}, pages 873--881, 2015.

\bibitem{gu2020cascade}
Xiaodong Gu, Zhiwen Fan, Siyu Zhu, Zuozhuo Dai, Feitong Tan, and Ping Tan.
\newblock Cascade cost volume for high-resolution multi-view stereo and stereo
  matching.
\newblock In {\em Proceedings of the IEEE Conference on Computer Vision and
  Pattern Recognition (CVPR)}, pages 2495--2504, 2020.

\bibitem{hosni2012fast}
Asmaa Hosni, Christoph Rhemann, Michael Bleyer, Carsten Rother, and Margrit
  Gelautz.
\newblock Fast cost-volume filtering for visual correspondence and beyond.
\newblock {\em IEEE Transactions on Pattern Analysis and Machine Intelligence
  (TPAMI)}, 35(2):504--511, 2012.

\bibitem{huang18deepmvs}
Po{-}Han Huang, Kevin Matzen, Johannes Kopf, Narendra Ahuja, and Jia{-}Bin
  Huang.
\newblock {DeepMVS}: Learning multi-view stereopsis.
\newblock In {\em Proceedings of the IEEE Conference on Computer Vision and
  Pattern Recognition (CVPR)}, pages 2821--2830, 2018.

\bibitem{im2019dpsnet}
Sunghoon Im, Hae{-}Gon Jeon, Stephen Lin, and In~So Kweon.
\newblock {DPSNet}: End-to-end deep plane sweep stereo.
\newblock In {\em International Conference on Learning Representations (ICLR)},
  2019.

\bibitem{ji2017surfacenet}
Mengqi Ji, Juergen Gall, Haitian Zheng, Yebin Liu, and Lu Fang.
\newblock Surfacenet: An end-to-end 3d neural network for multiview stereopsis.
\newblock In {\em Proceedings of the IEEE International Conference on Computer
  Vision (ICCV)}, pages 2307--2315, 2017.

\bibitem{kar2017learning}
Abhishek Kar, Christian H{\"a}ne, and Jitendra Malik.
\newblock Learning a multi-view stereo machine.
\newblock In {\em Advances in Neural Information Processing Systems (NeurIPS)},
  pages 364--375, 2017.

\bibitem{kendall2017uncertainties}
Alex Kendall and Yarin Gal.
\newblock What uncertainties do we need in bayesian deep learning for computer
  vision?
\newblock {\em Advances in Neural Information Processing Systems (NeurIPS)},
  2017.

\bibitem{DBLP:journals/corr/KingmaB14}
Diederik~P. Kingma and Jimmy Ba.
\newblock Adam: {A} method for stochastic optimization.
\newblock In Yoshua Bengio and Yann LeCun, editors, {\em International
  Conference on Learning Representations (ICLR)}, 2015.

\bibitem{kusupati2019normal}
Uday Kusupati, Shuo Cheng, Rui Chen, and Hao Su.
\newblock Normal assisted stereo depth estimation.
\newblock {\em Proceedings of the IEEE Conference on Computer Vision and
  Pattern Recognition (CVPR)}, 2020.

\bibitem{lee2019big}
Jin~Han Lee, Myung-Kyu Han, Dong~Wook Ko, and Il~Hong Suh.
\newblock From big to small: Multi-scale local planar guidance for monocular
  depth estimation.
\newblock {\em arXiv preprint arXiv:1907.10326}, 2019.

\bibitem{liang2018learning}
Zhengfa Liang, Yiliu Feng, Yulan Guo, Hengzhu Liu, Wei Chen, Linbo Qiao, Li
  Zhou, and Jianfeng Zhang.
\newblock Learning for disparity estimation through feature constancy.
\newblock In {\em Proceedings of the IEEE Conference on Computer Vision and
  Pattern Recognition (CVPR)}, pages 2811--2820, 2018.

\bibitem{long2020multiview}
Xiaoxiao Long, Lingjie Liu, Wei Li, Christian Theobalt, and Wenping Wang.
\newblock Multi-view depth estimation using epipolar spatio-temporal networks.
\newblock In {\em Proceedings of the IEEE Conference on Computer Vision and
  Pattern Recognition (CVPR)}, 2021.

\bibitem{luo2019p}
Keyang Luo, Tao Guan, Lili Ju, Haipeng Huang, and Yawei Luo.
\newblock {P}-{MVSNet}: Learning patch-wise matching confidence aggregation for
  multi-view stereo.
\newblock In {\em Proceedings of the IEEE International Conference on Computer
  Vision (ICCV)}, pages 10452--10461, 2019.

\bibitem{luo2020attention}
Keyang Luo, Tao Guan, Lili Ju, Yuesong Wang, Zhuo Chen, and Yawei Luo.
\newblock Attention-aware multi-view stereo.
\newblock In {\em Proceedings of the IEEE Conference on Computer Vision and
  Pattern Recognition (CVPR)}, pages 1590--1599, 2020.

\bibitem{mayer2016large}
Nikolaus Mayer, Eddy Ilg, Philip Hausser, Philipp Fischer, Daniel Cremers,
  Alexey Dosovitskiy, and Thomas Brox.
\newblock A large dataset to train convolutional networks for disparity,
  optical flow, and scene flow estimation.
\newblock In {\em Proceedings of the IEEE Conference on Computer Vision and
  Pattern Recognition (CVPR)}, pages 4040--4048, 2016.

\bibitem{murez2020atlas}
Zak Murez, Tarrence van As, James Bartolozzi, Ayan Sinha, Vijay Badrinarayanan,
  and Andrew Rabinovich.
\newblock Atlas: End-to-end {3D} scene reconstruction from posed images.
\newblock In {\em Proceedings of the European Conference on Computer Vision
  (ECCV)}, 2020.

\bibitem{nie2019multi}
Guang-Yu Nie, Ming-Ming Cheng, Yun Liu, Zhengfa Liang, Deng-Ping Fan, Yue Liu,
  and Yongtian Wang.
\newblock Multi-level context ultra-aggregation for stereo matching.
\newblock In {\em Proceedings of the IEEE Conference on Computer Vision and
  Pattern Recognition (CVPR)}, pages 3283--3291, 2019.

\bibitem{poms2018learning}
Alex Poms, Chenglei Wu, Shoou-I Yu, and Yaser Sheikh.
\newblock Learning patch reconstructability for accelerating multi-view stereo.
\newblock In {\em Proceedings of the IEEE Conference on Computer Vision and
  Pattern Recognition (CVPR)}, pages 3041--3050, 2018.

\bibitem{qi2018geonet}
Xiaojuan Qi, Renjie Liao, Zhengzhe Liu, Raquel Urtasun, and Jiaya Jia.
\newblock Geonet: Geometric neural network for joint depth and surface normal
  estimation.
\newblock In {\em Proceedings of the IEEE Conference on Computer Vision and
  Pattern Recognition (CVPR)}, pages 283--291, 2018.

\bibitem{ramachandran2019stand}
Prajit Ramachandran, Niki Parmar, Ashish Vaswani, Irwan Bello, Anselm Levskaya,
  and Jonathon Shlens.
\newblock Stand-alone self-attention in vision models.
\newblock {\em Advances in Neural Information Processing Systems (NeurIPS)},
  2019.

\bibitem{sfm16cvpr}
Johannes~L. Sch{\"{o}}nberger and Jan{-}Michael Frahm.
\newblock Structure-from-motion revisited.
\newblock In {\em Proceedings of the IEEE Conference on Computer Vision and
  Pattern Recognition (CVPR)}, pages 4104--4113, 2016.

\bibitem{shaw2018self}
Peter Shaw, Jakob Uszkoreit, and Ashish Vaswani.
\newblock Self-attention with relative position representations.
\newblock In {\em Proceedings of the 2018 Conference of the North American
  Chapter of the Association for Computational Linguistics: Human Language
  Technologies, Volume 2 (Short Papers)}, pages 464--468, 2018.

\bibitem{sturm12iros}
J. Sturm, N. Engelhard, F. Endres, W. Burgard, and D. Cremers.
\newblock A benchmark for the evaluation of {RGB-D} {SLAM} systems.
\newblock In {\em IEEE/RSJ International Conference on Intelligent Robots and
  Systems (IROS)}, 2012.

\bibitem{sun2018pwc}
Deqing Sun, Xiaodong Yang, Ming-Yu Liu, and Jan Kautz.
\newblock {PWC-Net}: {CNN}s for optical flow using pyramid, warping, and cost
  volume.
\newblock In {\em Proceedings of the IEEE Conference on Computer Vision and
  Pattern Recognition (CVPR)}, pages 8934--8943, 2018.

\bibitem{tankovich2020hitnet}
Vladimir Tankovich, Christian H{\"a}ne, Sean Fanello, Yinda Zhang, Shahram
  Izadi, and Sofien Bouaziz.
\newblock {HITNet}: Hierarchical iterative tile refinement network for
  real-time stereo matching.
\newblock {\em arXiv preprint arXiv:2007.12140}, 2020.

\bibitem{tobin2019geometry}
Josh Tobin, OpenAI Robotics, and Pieter Abbeel.
\newblock Geometry-aware neural rendering.
\newblock {\em Advances in Neural Information Processing Systems (NeurIPS)},
  2019.

\bibitem{tola2012efficient}
Engin Tola, Christoph Strecha, and Pascal Fua.
\newblock Efficient large-scale multi-view stereo for ultra high-resolution
  image sets.
\newblock {\em Machine Vision and Applications}, 23(5):903--920, 2012.

\bibitem{tung2019learning}
Hsiao-Yu~Fish Tung, Ricson Cheng, and Katerina Fragkiadaki.
\newblock Learning spatial common sense with geometry-aware recurrent networks.
\newblock In {\em Proceedings of the IEEE Conference on Computer Vision and
  Pattern Recognition (CVPR)}, pages 2595--2603, 2019.

\bibitem{ummenhofer17demon}
Benjamin Ummenhofer, Huizhong Zhou, Jonas Uhrig, Nikolaus Mayer, Eddy Ilg,
  Alexey Dosovitskiy, and Thomas Brox.
\newblock Demon: Depth and motion network for learning monocular stereo.
\newblock In {\em Proceedings of the IEEE Conference on Computer Vision and
  Pattern Recognition (CVPR)}, pages 5622--5631, 2017.

\bibitem{vaswani2017attention}
Ashish Vaswani, Noam Shazeer, Niki Parmar, Jakob Uszkoreit, Llion Jones,
  Aidan~N Gomez, Lukasz Kaiser, and Illia Polosukhin.
\newblock Attention is all you need.
\newblock {\em Advances in Neural Information Processing Systems (NeurIPS)},
  2017.

\bibitem{wang2021patchmatchnet}
Fangjinhua Wang, Silvano Galliani, Christoph Vogel, Pablo Speciale, and Marc
  Pollefeys.
\newblock Patchmatchnet: Learned multi-view patchmatch stereo.
\newblock In {\em Proceedings of the IEEE Conference on Computer Vision and
  Pattern Recognition (CVPR)}, pages 14194--14203, 2021.

\bibitem{wang2018non}
Xiaolong Wang, Ross Girshick, Abhinav Gupta, and Kaiming He.
\newblock Non-local neural networks.
\newblock In {\em Proceedings of the IEEE Conference on Computer Vision and
  Pattern Recognition (CVPR)}, pages 7794--7803, 2018.

\bibitem{wang2020mesh}
Yuesong Wang, Tao Guan, Zhuo Chen, Yawei Luo, Keyang Luo, and Lili Ju.
\newblock Mesh-guided multi-view stereo with pyramid architecture.
\newblock In {\em Proceedings of the IEEE Conference on Computer Vision and
  Pattern Recognition (CVPR)}, pages 2039--2048, 2020.

\bibitem{xiao2013sun3d}
Jianxiong Xiao, Andrew Owens, and Antonio Torralba.
\newblock {SUN3D}: A database of big spaces reconstructed using sfm and object
  labels.
\newblock In {\em Proceedings of the IEEE International Conference on Computer
  Vision (ICCV)}, pages 1625--1632, 2013.

\bibitem{xu2017multi}
Dan Xu, Elisa Ricci, Wanli Ouyang, Xiaogang Wang, and Nicu Sebe.
\newblock Multi-scale continuous {CRF}s as sequential deep networks for
  monocular depth estimation.
\newblock In {\em Proceedings of the IEEE Conference on Computer Vision and
  Pattern Recognition (CVPR)}, 2017.

\bibitem{xu2020aanet}
Haofei Xu and Juyong Zhang.
\newblock {AANet}: Adaptive aggregation network for efficient stereo matching.
\newblock In {\em Proceedings of the IEEE Conference on Computer Vision and
  Pattern Recognition (CVPR)}, pages 1959--1968, 2020.

\bibitem{xu2020learning}
Qingshan Xu and Wenbing Tao.
\newblock Learning inverse depth regression for multi-view stereo with
  correlation cost volume.
\newblock In {\em AAAI Conference on Artificial Intelligence (AAAI)}, pages
  12508--12515, 2020.

\bibitem{yang2020cost}
Jiayu Yang, Wei Mao, Jose~M Alvarez, and Miaomiao Liu.
\newblock Cost volume pyramid based depth inference for multi-view stereo.
\newblock In {\em Proceedings of the IEEE Conference on Computer Vision and
  Pattern Recognition (CVPR)}, pages 4877--4886, 2020.

\bibitem{Yang_2021_CVPR}
Zhenpei Yang, Li~Erran Li, and Qixing Huang.
\newblock Strumononet: Structure-aware monocular 3d prediction.
\newblock In {\em Proceedings of the IEEE Conference on Computer Vision and
  Pattern Recognition (CVPR)}, 2021.

\bibitem{yao2018mvsnet}
Yao Yao, Zixin Luo, Shiwei Li, Tian Fang, and Long Quan.
\newblock Mvsnet: Depth inference for unstructured multi-view stereo.
\newblock In {\em Proceedings of the European Conference on Computer Vision
  (ECCV)}, pages 767--783, 2018.

\bibitem{yao2019recurrent}
Yao Yao, Zixin Luo, Shiwei Li, Tianwei Shen, Tian Fang, and Long Quan.
\newblock Recurrent {MVSNet} for high-resolution multi-view stereo depth
  inference.
\newblock In {\em Proceedings of the IEEE Conference on Computer Vision and
  Pattern Recognition (CVPR)}, pages 5525--5534, 2019.

\bibitem{yao2020blendedmvs}
Yao Yao, Zixin Luo, Shiwei Li, Jingyang Zhang, Yufan Ren, Lei Zhou, Tian Fang,
  and Long Quan.
\newblock Blendedmvs: A large-scale dataset for generalized multi-view stereo
  networks.
\newblock {\em Proceedings of the IEEE Conference on Computer Vision and
  Pattern Recognition (CVPR)}, 2020.

\bibitem{yin2019enforcing}
Wei Yin, Yifan Liu, Chunhua Shen, and Youliang Yan.
\newblock Enforcing geometric constraints of virtual normal for depth
  prediction.
\newblock In {\em Proceedings of the IEEE International Conference on Computer
  Vision (ICCV)}, pages 5684--5693, 2019.

\bibitem{yu2020fast}
Zehao Yu and Shenghua Gao.
\newblock {Fast-MVSNet}: Sparse-to-dense multi-view stereo with learned
  propagation and gauss-newton refinement.
\newblock In {\em Proceedings of the IEEE Conference on Computer Vision and
  Pattern Recognition (CVPR)}, pages 1949--1958, 2020.

\bibitem{zbontar2015computing}
Jure Zbontar and Yann LeCun.
\newblock Computing the stereo matching cost with a convolutional neural
  network.
\newblock In {\em Proceedings of the IEEE Conference on Computer Vision and
  Pattern Recognition (CVPR)}, pages 1592--1599, 2015.

\bibitem{zhang2021long}
Xudong Zhang, Yutao Hu, Haochen Wang, Xianbin Cao, and Baochang Zhang.
\newblock Long-range attention network for multi-view stereo.
\newblock In {\em IEEE Winter Conference on Applications of Computer Vision
  (WACV)}, pages 3782--3791, 2021.

\end{thebibliography}
